\documentclass{article}

\PassOptionsToPackage{numbers, compress}{natbib}

\usepackage[main,final]{neurips_2026}

\usepackage[utf8]{inputenc}
\usepackage[T1]{fontenc}
\PassOptionsToPackage{hyphens}{url}
\usepackage{hyperref}
\usepackage{url}
\usepackage{booktabs}
\usepackage{amsfonts}
\usepackage{nicefrac}
\usepackage{microtype}
\usepackage{xcolor}

\usepackage[english]{babel}
\usepackage{amsmath}
\usepackage{amssymb}
\usepackage{amsthm}
\usepackage{bm}
\usepackage{algorithm}
\usepackage{algorithmic}
\usepackage{graphicx}
\usepackage{textcomp}

\usepackage{caption}
\usepackage{subcaption}

\usepackage[capitalize,noabbrev]{cleveref}

\usepackage{csquotes}

\theoremstyle{plain}
\newtheorem{theorem}{Theorem}[section]
\newtheorem{proposition}[theorem]{Proposition}
\newtheorem{lemma}[theorem]{Lemma}

\theoremstyle{definition}
\newtheorem{definition}[theorem]{Definition}

\theoremstyle{remark}
\newtheorem{remark}[theorem]{Remark}

\AddToHook{env/definition/begin}{\crefalias{theorem}{definition}}
\AddToHook{env/proposition/begin}{\crefalias{theorem}{proposition}}
\AddToHook{env/lemma/begin}{\crefalias{theorem}{lemma}}
\AddToHook{env/remark/begin}{\crefalias{theorem}{remark}}

\crefname{theorem}{Theorem}{Theorems}
\crefname{definition}{Definition}{Definitions}
\crefname{proposition}{Proposition}{Propositions}
\crefname{lemma}{Lemma}{Lemmas}
\crefname{remark}{Remark}{Remarks}

\DeclareMathOperator{\ar}{ar}
\DeclareMathOperator{\diag}{diag}
\DeclareMathOperator{\interior}{int}
\DeclareMathOperator{\median}{median}
\DeclareMathOperator*{\argmax}{arg\,max}
\DeclareMathOperator*{\argmin}{arg\,min}

\usepackage{tabularx}

\usepackage{placeins}

\usepackage{siunitx}
\DeclareSIUnit\knot{kn}
\DeclareSIUnit\nauticalmile{NM}
\DeclareSIUnit\foot{ft}
\DeclareSIUnit\flops{FLOPS}

\usepackage{tabularray}
\UseTblrLibrary{booktabs,siunitx}

\usepackage{bbding}
\usepackage{tikz}
\usepackage[edges]{forest}
\usepackage{wheelchart}
\usetikzlibrary{
    angles,
    arrows,
    arrows.meta,
    backgrounds,
    bbox,
    calc,
    decorations.pathreplacing,
    decorations.text,
    fit,
    intersections,
    math,
    patterns,
    positioning,
    quotes,
    shapes.geometric
}
\tikzset{
    compass/.pic = {
            \filldraw[pic actions,rotate=90,scale=0.4] (0,0) -- (135:1) -- (0:2) node[transform shape,rotate=-90,above,scale=3]{N};
            \filldraw[pic actions,rotate=90,scale=0.4,fill=white] (0,0) -- (-135:1) -- (0:2) -- cycle;
        }
}

\usepackage{pgfplots}
\usepackage{pgfplotstable}
\pgfplotsset{compat=newest}
\usepgfplotslibrary{%
    fillbetween,
    groupplots,
    statistics
}
\pgfplotsset{
    discard if not/.style 2 args={
        x filter/.append code={
            \edef\tempa{\thisrow{#1}}
            \edef\tempb{#2}
            \ifx\tempa\tempb
            \else
                
            \fi
        }
    },
    autosafe plot/.style={
        width=0.8684\linewidth,
        height=0.8684\linewidth,
        scale only axis,
        scaled ticks=false,
        tick pos=left,
        tick align=outside,
        tick label style={/pgf/number format/fixed},
        legend cell align=left,
        every axis plot/.append style={mark size=1.7pt,line width=1pt},
    },
    autosafe heatmap/.style={
        autosafe plot,
        view={0}{90},
        colormap/viridis,
    },
    autosafe sensitivity heatmap/.style={
        autosafe heatmap,
        xlabel={Width multiple \(s\)},
        ylabel={Decay \(\gamma{}\)},
        xmin=-0.5,
        xmax=3.5,
        ymin=-0.5,
        ymax=2.5,
        xtick={0,1,2,3},
        xticklabels={1,2,3,5},
        ytick={0,1,2},
        yticklabels={0.5,1,2},
    },
    autosafe sensitivity cells/.style={
        matrix plot*,
        mesh/cols=4,
        point meta=explicit,
    },
}

\newcommand{\autosafepairedplotbounds}[4]{%
    \pgfresetboundingbox
    \path[use as bounding box]
        ([xshift=-#1,yshift=-#3]current axis.south west)
        rectangle
        ([xshift=#2,yshift=#4]current axis.north east);
}

\definecolor{DLRBlack}{gray}{0}
\definecolor{DLRGrey}{gray}{0.420} 
\colorlet{DLRGray}{DLRGrey}
\definecolor{DLRWhite}{gray}{1}

\colorlet{DLREagleColor}{DLRBlack}
\colorlet{DLRTextColor}{DLRBlack}

\colorlet{DLRDarkerGrey}{DLRGrey}
\definecolor{DLRDarkGrey}{gray}{0.537} 
\definecolor{DLRMediumGrey}{gray}{0.702} 
\definecolor{DLRLightGrey}{gray}{0.820} 
\definecolor{DLRLighterGrey}{gray}{0.929} 

\colorlet{DLRDarkerGray}{DLRDarkerGrey}
\colorlet{DLRDarkGray}{DLRDarkGrey}
\colorlet{DLRMediumGray}{DLRMediumGrey}
\colorlet{DLRLightGray}{DLRLightGrey}
\colorlet{DLRLighterGray}{DLRLighterGrey}

\definecolor{DLRDarkerBlue}{RGB}{0, 106, 144}
\definecolor{DLRDarkBlue}{RGB}{0, 156, 208}
\definecolor{DLRBlue}{RGB}{33, 187, 223}
\colorlet{DLRMediumBlue}{DLRBlue}
\definecolor{DLRLightBlue}{RGB}{149, 212, 238}
\definecolor{DLRLighterBlue}{RGB}{201, 232, 251}

\definecolor{DLRDarkerGreen}{RGB}{115, 163, 63}
\definecolor{DLRDarkGreen}{RGB}{158, 193, 76}
\definecolor{DLRGreen}{RGB}{199, 214, 84}
\colorlet{DLRMediumGreen}{DLRGreen}
\definecolor{DLRLightGreen}{RGB}{215, 223, 116}
\definecolor{DLRLighterGreen}{RGB}{228, 234, 173}

\definecolor{DLRDarkerYellow}{RGB}{224, 177, 57}
\definecolor{DLRDarkYellow}{RGB}{254, 206, 73}
\definecolor{DLRYellow}{RGB}{255, 223, 73}
\colorlet{DLRMediumYellow}{DLRYellow}
\definecolor{DLRLightYellow}{RGB}{255, 234, 117}
\definecolor{DLRLighterYellow}{RGB}{255, 248, 189}

\definecolor{DLRDarkestBlue}{RGB}{0, 50, 69}
\definecolor{DLRDarkestGreen}{RGB}{99, 119, 34}
\definecolor{DLRDarkestYellow}{RGB}{190, 150, 0}

\definecolor{DLRRed}{RGB}{179, 63, 61}
\colorlet{DLRMediumRed}{DLRRed}
\definecolor{DLRLightRed}{RGB}{199, 122, 109}
\definecolor{DLRLighterRed}{RGB}{223, 180, 168}

\definecolor{DLRDarkestGray}{gray}{0.302} 

\title{Defining Operational Conditions for Safety-Critical AI-Based Systems from Data}

\author{%
  Johann M.~Christensen\\
  Institute for AI Safety and Security\\
  German Aerospace Center (DLR)\\
  Sankt Augustin, Germany\\
  \texttt{johann.christensen@dlr.de}\\
  \And{}
  Elena Hoemann\\
  Institute for AI Safety and Security\\
  German Aerospace Center (DLR)\\
  Sankt Augustin, Germany\\
  \texttt{elena.hoemann@dlr.de}\\
  \And{}
  Frank Köster\\
  Institute for AI Safety and Security\\
  German Aerospace Center (DLR)\\
  Sankt Augustin, Germany\\
  \texttt{frank.koester@dlr.de}\\
  \And{}
  Sven Hallerbach\\
  Institute for AI Safety and Security\\
  German Aerospace Center (DLR)\\
  Sankt Augustin, Germany\\
  \texttt{sven.hallerbach@dlr.de}%
}

\begin{document}

\maketitle{}

\begin{abstract}
    Artificial Intelligence (AI) has been on the rise in many domains, including numerous safety-critical applications.
    However, for complex systems in the real world, defining the underlying environmental conditions in which the AI-based system must operate---the Operational Design Domain (ODD)---is extremely challenging.
    This often results in an incomplete description of the ODD, which contrasts with the requirements of many domains for certifying AI-based systems.
    Traditionally, the ODD is created in the early stages of the development process, drawing on sophisticated expert knowledge and related standards.
    This paper presents a novel method for defining the ODD a posteriori from previously collected data using a multidimensional kernel-based representation.
    This approach is validated through both synthetic benchmarks and a real-world aviation use case.
    Moreover, the paper defines similarity of two ODDs if they generate the same outputs up to Lebesgue-null input sets and proves convergence in volume of the calibrated representation under the stated assumptions.
    The novel, safety-by-design, deterministic kernel-based ODD representation is derived fully automatically, given documented assurance inputs, permutation-stable, bounded by construction, and, under affine-equivariant per-dimension normalization, invariant to the choice of units.
    Utilizing the proposed ODD representation supports future certification of data-driven, safety-critical AI-based systems.
\end{abstract}

\section{Introduction}\label{sec:Introduction}
The rapid adoption of Artificial Intelligence (AI) across a wide range of application domains has fundamentally changed how complex systems are designed and operated.
In safety-critical domains, AI-based systems are increasingly responsible for perception, decision-making, and control~\cite{Constans2007, Julian2019a, Ziakkas2023, Bello2024, Erdogan2024, GreinerFuchs2024, Mandal2024, Stefani2024b, Schlichting2025}.
While these capabilities enable unprecedented levels of autonomy and performance, they also introduce new challenges for safety assurance, validation, and certification~\cite{Forsberg2020, Tambon2022}.
Research into the safety of AI-based systems has only recently begun to catch up with their practical deployment.
Especially in domains with stringent safety requirements, e.g., aviation, safety considerations are not merely desirable but mandatory to enable certification and operational approval~\cite{Mamalet2021, Katz2023a, Rabe2021, Christensen2025}.
Current approaches to ensuring the safety of AI-based systems predominantly rely on reactive measures, such as extensive testing, runtime monitoring, and post hoc failure analysis~\cite{Colwell2018, Sun2022a, Torens2024}.
Although effective, these approaches are often time-consuming, costly, and difficult to scale.
In contrast, safety-by-design methodologies aim to integrate safety considerations directly into the system development process, favoring proactive design measures over reactive mitigation.
This paradigm shift has given rise to the research field of AI engineering, which seeks systematic, certifiable methods for designing AI-based systems from the outset.
A central concept in safety-by-design AI engineering is the Operational Design Domain (ODD)~\cite{Colwell2018, BSI2020, Werner2026}.
The ODD specifies the set of operational conditions under which an AI-based system is intended to function safely~\cite{BSI2020, SAE-J3016, ISO34503, EUASA2024}.
By explicitly defining these conditions, the ODD provides a foundation for system design and verification, and has become a key instrument in multiple regulatory frameworks.
However, despite its importance, defining an ODD remains challenging~\cite{Gyllenhammar2020, Cappi2024}.
Traditionally, ODDs are created manually by domain experts early in the development process~\cite{Gyllenhammar2020, ISO34503, Stefani2023}.
While expert-defined ODDs are effective when system behavior and environmental conditions are well understood, they become increasingly difficult to construct for complex real-world systems.
Prior research characterizes ODDs from data in an aeronautical context~\cite{Kaakai2023}; however, the approach provides neither a deterministic, order-independent representation nor a formal framework for ODD similarity---both of which support the certification use cases addressed in this work (cf.\ \cref{sec:EASAObjectives}).
Data-driven ODDs promise several advantages: they can capture complex, implicit parameter dependencies, naturally reflect real operational conditions, and be updated as new data become available.
This work addresses this gap by introducing a deterministic, kernel-based framework for data-driven ODD construction.
Instead of manually creating an ODD and iteratively optimizing it, the proposed method constructs the ODD directly from data using samples as anchor points for analytically defined kernel functions.
The resulting ODD representation is continuous and bounded and, for a fixed calibration procedure, permutation-stable and uniquely determined by the available data.
Importantly, the method remains applicable even in sparse-data regimes, making it suitable for both early development stages and mature systems.
This work aims to provide a mathematically rigorous framework for deriving ODDs solely from data and supporting safety-by-design assurance activities.
This work formalizes ODDs as mathematical structures, introduces a kernel-based affinity representation, and proposes a procedure for parameter selection and handling of out-of-distribution samples that runs fully automatically, given documented assurance inputs.
By construction, the resulting ODD representation is bounded, deterministic, and interpretable, enabling its use not only for modeling and monitoring but also as a foundation for future certification of AI-based systems.

\section{State of the Art}\label{sec:SoA}
The ODD concept has gained acceptance in the automotive and aviation domains, with other domains now following~\cite{GreinerFuchs2024, Huseynzade2023, Adedjouma2024, Stefani2025}.
ODDs are intended to represent the real-world conditions under which automated systems, including AI-based systems, are expected to function, but necessarily remain approximations of those conditions~\cite{BSI2020, SAE-J3016, ISO34503, EUASA2024}.
Originating from the automotive domain, the ODD is loosely defined as the specific conditions, such as environmental, geographic, and time-of-day restrictions, under which an autonomous driving system is expected to function~\cite{SAE-J3016, ISO34503}.
The definition of the European Union Aviation Safety Agency (EASA) is similar, extending it to any AI-based system and, by extension, to any artificial intelligence/machine learning (AI/ML) constituent~\cite{EUASA2024}.
Moreover, the ODD is split into two parts: the taxonomy and the ontology~\cite{BSI2020, Trypuz2024}.
The taxonomy describes the overall set of parameters and their applicable ranges, e.g., the car's current speed or the aircraft's altitude.
The ontology, however, describes the interrelations among those parameters and how certain parameters influence one another.
An example here might be the minimum possible turning radius of a car, depending on its speed, or the landing speed of an aircraft, depending on current weather and visibility.
As such, the ODD plays a critical role in safety-by-design AI engineering methodologies to ensure compliance with applicable standards and regulations.
However, creating the ODD is not trivial.
While the ranges of some parameters can be quite evident, others, especially their interconnections, are almost impossible to define beforehand, for example, if they depend on complex processes like weather.

This clear commitment to ODDs as a standardized framework for describing operational conditions requires a comprehensive representation used not only for modeling and information exchange but also for monitors that can detect whether the system still operates within the ODD~\cite{BSI2020, EUASA2024, Cappi2024}.
Although some ODD representations use structured representations such as tables or ASAM's YAML-based OpenODD format~\cite{Werner2026, VFSAM2021, VFSAM2025}, extensive mathematical representations remain a subject of ongoing research.
Most works represent the ODD as an \(n\)-dimensional space, with different approaches to constraining it, e.g., convex polytopes, which provide a clean representation but cannot model nonlinear ODD ontologies~\cite{Nenchev2025}.
Others provide an extensive framework for describing the ODD in great detail, but the ontology representation cannot handle constraints imposed by parameter combinations~\cite{Shakeri2024}.
Research has also been conducted to split the ODD into \emph{micro ODDs} (\(\mu{}\)ODDs), which are used to improve handling and cover potentially complex ODDs~\cite{Koopman2019a, Salvi2022}.
Thus, while ODDs can contain abstract definitions of the environment, they are, in general, currently defined by a parameter space that describes the overall taxonomy.

Constructing a membership region from sampled in-distribution (ID) points also connects the ODD-from-data problem to one-class classification and statistical support estimation.
Classical estimators such as the one-class support vector machine~\cite{Schoelkopf2001}, support vector data description~\cite{Tax2004}, and kernel-density superlevel sets~\cite{Devroye1980} likewise enclose a sampled distribution.
The proposed affinity is globally a noisy-OR aggregation and, when overlap terms are small, is closely related to a variable-bandwidth kernel-density estimate through the per-anchor widths of \cref{eq:sigma_kk_mod}; scalable density-matrix approximations to kernel-density estimation also exist~\cite{Gonzalez2022, Gallego2022}.
Unlike the other estimators, the construction in this work targets the ODD-specific combination of a deterministic, permutation-stable membership region, an explicit out-of-distribution (OOD) consistency constraint with a proven termination bound, and a bounded affinity suitable for formal safety argumentation.
Neural OOD detectors are excluded because incorporating learned detectors would introduce a separate training and assurance problem---which, in a safety-by-design setting, would itself require another ODD---beyond the scope of this work.
The quantitative comparison with representative one-class baselines is summarized in \cref{sec:AdditionalValidation}.

From the ODD, scenarios can be derived that provide a high-level description of operating conditions defined through a subset of parameter ranges~\cite{Ulbrich2015, Weissensteiner2023}.
For example, the scenarios driving and flying at night in rainy weather would define the time of day to be between 6 p.m.\ and 6 a.m.\ and the precipitation and cloud coverage to be greater than zero.
Sometimes, this is further split into logical and concrete scenarios, where logical scenarios define the aforementioned ranges, while concrete scenarios specify initial values for each parameter~\cite{Zhang2024}.
Nevertheless, a scenario is always a series of snapshots, called scenes~\cite{Ulbrich2015}.
Each scene is therefore a unique set of ODD parameter values, together with data from all sensors, given the current environmental conditions.

\section{Mathematical Representation of ODDs}\label{sec:MathODD}
As previously described, multiple frameworks exist for structuring the ODD\@.
However, all those representations lack a comprehensive yet generic framework for representing the ontology, as well as a means to compare two ODDs and determine whether they express the same problem.
Both are solved with the following mathematical representation of an ODD\@.

\begin{definition}[ODD Structures]\label{def:ODDStructure}
    Let the ODD be defined as the structure \(\mathcal{O} = (X, R_1^\mathcal{O}, \dots, R_r^\mathcal{O}, f^\mathcal{O}, \Omega^\mathcal{O})\), where \(X \subseteq \mathbb{R}^n\) is the domain, also called taxonomy, \(R_i^\mathcal{O} = \{\bm{x} \in X \mid R_i(\bm{x}) = 1\} \subseteq X\) is the set of points satisfying the \(i\)-th predicate, also called ontology, and \(f^\mathcal{O}\colon X \times \Omega^\mathcal{O} \to Y\), with the set of admissible noise values \(\Omega^\mathcal{O}\) and the codomain \(Y\), is the (stochastic) interpretation function.
    Thus, the signature is defined as \(\sigma = \{R_1, \dots, R_r, f, \Omega{}\}{}\) and the arities as \(\ar(R_i) = 1\), \(\ar(f) = n + \dim(\Omega)\), and \(\ar(\Omega) = 0\).
\end{definition}
\begin{definition}[Similarity of ODD Structures]\label{def:ODDSimilarity}
    Let \(X_j \subseteq \mathbb{R}^{n_j}\) be the taxonomy of \(\mathcal{O}_j\), \(j \in \{1, 2\}{}\).
    Let \(\mathcal{L}^m\) denote the \(m\)-dimensional Lebesgue measure.
    Two ODD structures \(\mathcal{O}_1\) and \(\mathcal{O}_2\) are considered similar, written \(\mathcal{O}_1 \sim \mathcal{O}_2\), if and only if their generated datasets agree after discarding Lebesgue-null subsets of their input regions:
    \begin{align}
        \mathcal{O}_1 \sim \mathcal{O}_2
        \stackrel{\mathrm{def}}{\iff}
         & \exists Z_1 \subseteq \mathcal{R}^{\mathcal{O}_1}, Z_2 \subseteq \mathcal{R}^{\mathcal{O}_2}\colon
        \mathcal{L}^{n_1}(Z_1) = \mathcal{L}^{n_2}(Z_2) = 0,
        \nonumber                                                                                                  \\
         & f^{\mathcal{O}_1}\left((\mathcal{R}^{\mathcal{O}_1} \setminus Z_1) \times \Omega^{\mathcal{O}_1}\right)
        =
        f^{\mathcal{O}_2}\left((\mathcal{R}^{\mathcal{O}_2} \setminus Z_2) \times \Omega^{\mathcal{O}_2}\right)
        \text{.}
    \end{align}
    Here, \(\mathcal{R}^{\mathcal{O}}\) is the conjunction of all relational predicates in \(\mathcal{O}\), defined as \(\mathcal{R}^{\mathcal{O}} \equiv \bigcap_{i = 1}^r R_i^\mathcal{O}\).
\end{definition}
This image-based notion is data-centric and intentionally abstracts away semantic differences, output frequencies, and safety weighting; refinements that compare induced output distributions or restrict comparison to hazardous outputs are left to future work.\footnote{%
    As an example, given the definition of similarity, let two ODDs be defined as
    \(\mathcal{O}_1 = (\mathbb{R}^2, \{\bm{x} \in X \mid x_1 \geq 0\}, \{\bm{x} \in X \mid x_2 \geq 0\}, \{\bm{x} \in X \mid x_1 + x_2 \leq 2\}, f^{\mathcal{O}_1}(\bm{x}) = x_1 + x_2, \Omega^{\mathcal{O}_1} = \{0\})\)
    and
    \(\mathcal{O}_2 = (\{x \in \mathbb{R} \mid 0 \leq x \leq 2\}, f^{\mathcal{O}_2}(x) = x, \Omega^{\mathcal{O}_2} = \{0\})\).
    Thus, both generated output images equal \([0, 2]\), and therefore, \(\mathcal{O}_1 \sim \mathcal{O}_2\).
}

\section{The Kernel-Based Representation}\label{sec:Kernel}
Polytope-based ODD representations~\cite{Nenchev2025} (cf.\ \cref{sec:ODD_Representation}) work well for expert-defined ODDs but are unsuitable for data-driven construction: convex hulls may enclose regions outside the true ODD, and neural network--based representations are not order-independent, introducing
avoidable uncertainty~\cite{Kochenderfer2026}.
Instead, the proposed method assigns a kernel function to each anchor point and derives the ODD directly from data, without expert knowledge.
The resulting representation is continuous and bounded, permutation-stable, and uniquely determined by the data under the calibration procedure of \cref{subsec:KernelParameters}---in both sparse and dense regimes.

\subsection{Formal Definition of the Kernel-Based ODD}\label{subsec:FormalDefinition}
The following subsection introduces a formal definition, followed by parameterization and algorithmic construction of the data-driven ODD\@.
Let \(X \subseteq \mathbb{R}^n\) denote the parameter space of the ODD, where each point \(\bm{x} \in X\) represents a concrete operational condition described by \(n\) continuous parameters.
\begin{definition}[Samples and Anchor Points]\label{def:SamplesAnchorPoints}
    Let \(\mathcal{D} = \mathcal{D}_{\mathrm{ID}} \cup \mathcal{D}_{\mathrm{OOD}} \subseteq X\) be a finite dataset, where
    \(\mathcal{D}_{\mathrm{ID}} \subseteq X\) denotes the set of in-distribution (ID) samples that are considered part of the ODD, and \(\mathcal{D}_{\mathrm{OOD}} \subset X\) denotes the set of out-of-distribution (OOD) samples that are explicitly not part of the ODD\@.
    Thus, \(\mathcal{D}_\mathrm{ID} \cap \mathcal{D}_\mathrm{OOD} = \emptyset{}\).
    Before anchor selection, the ID input records are de-duplicated using resolution cells fixed by documented per-dimension acquisition resolutions.
    Each occupied cell retains the observed record nearest its mean; the mean is accumulated over lexicographically ordered records, and ties are broken lexicographically.
    Merged record identifiers and visit counts are retained as provenance and coverage evidence but do not affect the ODD geometry.
    Cross-class conflicts must be resolved by the surrounding data-management process so that \(\mathcal{D}_{\mathrm{ID}} \cap \mathcal{D}_{\mathrm{OOD}} = \emptyset{}\); the OOD samples themselves are not merged, and all remain subject to the consistency constraint of \cref{def:OODConsCons}.
    The anchor set \(\mathcal{A} \subseteq \mathcal{D}_{\mathrm{ID}}\) is the finite (sub)set of retained ID samples used to parameterize the kernel-based ODD representation.
    Retained anchors and OOD samples are indexed in lexicographic order; geometrically identical OOD records are interchangeable because their provenance does not affect the construction.
    In general, all retained ID samples are used, i.e., \(\mathcal{A} \equiv \mathcal{D}_{\mathrm{ID}}\).
\end{definition}

\begin{definition}[Local Affinity Function]\label{def:LocalAffinity}
    For each anchor point \(\bm{x}_i \in \mathcal{A}\), \(i = 1, \dots, N\), a local affinity function \(\alpha_i\colon X \to [0, 1]\) is defined using a positive-definite kernel.
\end{definition}

\begin{definition}[Global ODD Affinity Function]\label{def:GlobalAffinity}
    The kernel-based ODD representation is defined as a global affinity function \(\alpha\colon X \to [0, 1]\) constructed by superposition of all local affinity functions:
    \begin{align}
        \alpha(\bm{x}) = 1 - \prod_{i = 1}^{N} \left(1 - \alpha_i(\bm{x})\right)\text{.}\label{eq:Superposition}
    \end{align}
    The value \(\alpha(\bm{x})\) represents the degree to which the operational condition \(\bm{x}\) belongs to the ODD\@.
    The superposition is exactly the noisy-OR aggregation of the local affinities.
    Its deviation from their sum is bounded by
    \begin{align}
        0 \leq \sum_{i = 1}^{N} \alpha_i(\bm{x}) - \alpha(\bm{x}) \leq \sum_{i < j} \alpha_i(\bm{x}) \alpha_j(\bm{x})\text{.} \label{eq:NoisyORAdditiveBound}
    \end{align}
\end{definition}

\begin{definition}[Threshold-Based ODD Membership]
    Given the global affinity function \(\alpha{}\), let \(\zeta \in (0, 1)\) be the affinity threshold.
    For the deployed construction, \(\zeta{}\) is obtained by the split-conformal calibration of \cref{subsec:ConformalThreshold} from a held-out ID set and a documented false-exclusion budget \(\varepsilon{}\).
    A sample \(\bm{x} \in X\) is considered to be \emph{inside} the data-driven ODD if and only if
    \begin{align}
        \alpha(\bm{x}) \geq \zeta\text{.}
    \end{align}
\end{definition}

\begin{definition}[OOD Consistency Constraint]\label{def:OODConsCons}
    Assume \(\mathcal{A} \cap \mathcal{D}_{\mathrm{OOD}} = \emptyset{}\).
    Let \(\xi \in (0, \zeta)\), with \(\zeta{}\) the membership threshold of the data-driven ODD, denote a predefined maximum allowed affinity for OOD samples; \(\xi < \zeta{}\) guarantees that every observed OOD sample lies strictly outside the data-driven ODD\@.
    The kernel-based ODD representation shall satisfy
    \begin{align}
        \alpha(\bm{x}) \leq \xi \quad \forall \bm{x} \in \mathcal{D}_{\mathrm{OOD}}\text{.}
    \end{align}
    If this constraint is violated, the most contributing kernels must be adjusted, while keeping the anchor points fixed, until the constraint is satisfied.
\end{definition}
In practice, at each iteration, a globally most-violated OOD sample \(\bm{x}^* \in \argmax_{\bm{x} \in \mathcal{D}_\mathrm{OOD}} \alpha(\bm{x})\) is selected, with the lowest canonical OOD-sample index selected on an exact tie.
A kernel with the largest contribution to \(\alpha(\bm{x}^*)\), i.e., \(i^* \in \argmax_i \alpha_i(\bm{x}^*)\), is then identified, again with the lowest canonical anchor index selected on an exact tie, and its covariance matrix is scaled by a fixed factor \(c \in (0, 1)\).
This process is repeated until all OOD samples satisfy \(\alpha(\bm{x}) \leq \xi{}\).
Its termination is proven in \cref{sec:ProofofTermination}.
Together with canonical indexing and exact nearest-neighbor search, these tie rules make the construction permutation-stable in exact arithmetic.
Repeated runs on identical inputs return identical output.

The previous definitions, together with the notation introduced in \cref{sec:MathODD}, are now used to define the kernel-based ODD\@.
\begin{definition}[Kernel-based Operational Design Domain]\label{def:KernelODD}
    A kernel-based ODD is defined as the structure
    \begin{align}
        \mathcal{O} = (X, R^\mathcal{O}_\mathrm{M}, f^\mathcal{O}, \Omega^\mathcal{O}, a^\mathcal{O}_1, \dots, a^\mathcal{O}_N)\text{,}
    \end{align}
    where \(R^\mathcal{O}_\mathrm{M} = \{\bm{x} \in X \mid \alpha(\bm{x}) \geq \zeta{}\}{}\) is the sole membership predicate and \(a^\mathcal{O}_i \in \mathcal{A}\) are the anchor points with \(\ar(a_i^\mathcal{O}) = 0\).
\end{definition}

The fixed-parameter coverage and outer-halo behavior of the kernel-based representation, together with the two-sided convergence in volume of the calibrated construction, are analyzed in \cref{sec:AsympSimProof}.

\subsection{Choice of Kernel}
The affinity function must be continuous, bounded in \([0, 1]\), approach \num{0} for \(\|\bm{x} - \bm{x}_i\|_2 \to \infty{}\), and scale to arbitrarily many dimensions.
These requirements are satisfied by standard kernels such as the Laplacian kernel~\cite{Rupp2015}, the leptokurtic Cauchy--Lorentz kernel~\cite{Feller2009}, the cubic spline kernel~\cite{Monaghan1992}, and the radial basis function (RBF) kernel~\cite{Hastie2009}, used in this work. Its kernel function is defined as
\begin{align}
    \alpha_{i}(\bm{x}) = \exp\left(-\frac{1}{2}(\bm{x} - \bm{x}_{i})^\top \bm{\Sigma}_{i}^{-1}(\bm{x} - \bm{x}_{i})\right)\text{,}
\end{align}
for the \(i\)-th anchor point, where \(\bm{x}_{i} \in \mathbb{R}^n\) is the anchor point, and \(\bm{\Sigma}_{i} \in \mathbb{R}^{n \times n}\) is the symmetric, positive-definite free parameter matrix.
The multiplicative superposition in \cref{def:GlobalAffinity} ensures the global affinity remains bounded in \([0, 1]\) even when anchor points coincide.
Among the kernels above, the RBF kernel is selected for two reasons: its parameter matrix directly encodes per-dimension squared length scales, matching the nearest-neighbor parameterization of \cref{subsec:KernelParameters} and the diagonal approximation analyzed in \cref{sec:JustDiagSigma}, and its logarithm is a quadratic form, permitting the exact, numerically stable log-space evaluation of \cref{subsec:NumericalAffinity}.
The affinity superposition itself is kernel-agnostic; other kernels satisfying the stated requirements can be substituted, but their parameter calibration and termination analysis must be adapted accordingly.

\subsection{Choosing the Kernel Parameters}\label{subsec:KernelParameters}
Almost every kernel has a free parameter matrix.
Defining all those parameters by hand would undermine the purpose of an automated, data-driven ODD derivation framework.
Moreover, for the RBF kernel, \(N\) \(n\)-dimensional anchor points would lead to up to \(N n^2\) parameters to define.
To allow automated parameter definition, some assumptions and simplifications must be made.
First, to avoid implicit bias, the parameter ranges should be normalized.
This would ensure that all ranges are again of the same order of magnitude.
Thus, their corresponding matrix entries are also roughly of the same order of magnitude, reducing the likelihood that extreme matrix entries make the affinity function unstable.
Second, the coordinate components of nearby anchor points are assumed to be locally independent, while their relationships are governed only by global effects.
For the Gaussian RBF kernel, this assumption yields an axis-aligned factorization over the parameter dimensions and hence a diagonal \(\bm{\Sigma}_i\)~\cite{Bishop2006}.
This construction combines the local-independence assumption with the principle of local stationarity, under which a process may behave as stationary within sufficiently small regions although its structure varies globally~\cite{Fuentes2002}.
Under this assumption, the diagonal construction captures the relevant axis-aligned local scaling without introducing unsupported local cross-coordinate terms.
\Cref{sec:JustDiagSigma} gives an exact error bound relative to a corresponding hypothetical full-covariance kernel when the assumption is violated; the discrepancy depends on the omitted off-diagonal structure and need not vanish solely because anchor density increases.
The geometric density of distinct retained operational states provides evidence of local support, whereas repeated observations within the documented acquisition resolution provide coverage evidence only and are removed before constructing the geometry.
The corresponding parameter in the \(\bm{\Sigma}\) matrix should thus increase with the density of distinct nearby anchors, resulting in a wider local kernel influence.
Therefore, the parameters depend on nearest-neighbor distances between de-duplicated anchors; genuinely distinct nearby anchors still amplify one another through \cref{eq:Superposition}.
Thus, the entries of the \(\bm{\Sigma}\) matrix can be defined as
\begin{align}
    \sigma_{kk}^{(i)} = \kappa \exp(-\eta d_{ik}^*), \quad k = 1, \dots, n\text{,}
    \label{eq:sigma_kk}
\end{align}
where \(\sigma_{kk}^{(i)}\) are the diagonal entries of \(\bm{\Sigma}_i\), \(\kappa{}\) is the maximum diagonal entry, and \(\eta{}\) is the decay factor.
In global mode, \(j_i^* := \min \argmin_{j \ne i} \|\bm{x}_i - \bm{x}_j\|_2\), where the minimum selects the lowest canonical anchor index, and \(d_{ik}^* := \lvert x_{ik} - x_{j_i^* k}\rvert\); in per-dimension mode, \(d_{ik}^* := \min_{j \ne i} \lvert x_{ik} - x_{jk}\rvert\).
This approach reduces the number of manually defined parameters from \(N n^2\) to 2, or \(2n\) if \(\kappa{}\) and \(\eta{}\) are defined per dimension.

This, however, can lead to unstable matrix entries when anchor points are spread far apart, as the corresponding \(\sigma_{kk}^{(i)}\) will be rather small and, in turn, the corresponding entries in \(\bm{\Sigma}^{-1}\) can become numerically unstable.
Here, a possible approach is to define a lower limit for \cref{eq:sigma_kk}.
This would lead to
\begin{align}
    \sigma_{kk}^{(i)} = (\kappa - \lambda) \exp(-\eta d_{ik}^*) + \lambda, \quad k = 1, \dots, n\text{,}
    \label{eq:sigma_kk_mod}
\end{align}
where \(\lambda \ll \kappa{}\) is the lower bound of \(\sigma_{kk}^{(i)}\), specified relative to the maximum value, \(\lambda = \lambda_{\mathrm{rel}} \kappa{}\) with \(\lambda_{\mathrm{rel}} = \exp(-10)\), which bounds the condition number of \(\bm{\Sigma}^{-1}\) by \(1/\lambda_{\mathrm{rel}}\) and, unlike an absolute lower bound, preserves the scale invariance established below.

While \cref{eq:sigma_kk,eq:sigma_kk_mod} reduce the parameter count, the remaining choice of \(\kappa{}\) and \(\eta{}\) is crucial.
The exponent of \cref{eq:sigma_kk_mod} must be dimensionless and \(\sigma_{kk}^{(i)}\) is a variance, so \(\eta{}\) carries the unit of an inverse length and \(\kappa{}\) that of a squared length.
Fixing numerical values such as \(\kappa = \eta = 1\) therefore implicitly selects a unit of length, and the resulting ODD changes under the very normalization recommended above.
To resolve this, the parameters are calibrated from the data itself.
Let the full-space nearest-neighbor distances and their median be \(d_i^* := \min_{j \ne i} \|\bm{x}_i - \bm{x}_j\|_2\) and \(\tilde{d} := \median_i d_i^*\), respectively.
The calibrated parameters are defined as
\begin{align}
    \eta = \frac{\gamma}{\tilde{d}}, \qquad \kappa = \left(s \tilde{d}\right)^2, \qquad \lambda = \lambda_{\mathrm{rel}} \kappa\text{,} \label{eq:sigma_calibrated}
\end{align}
with dimensionless constants \(\gamma{}\) and \(s\), with defaults \(\gamma = 1\) and the scheduling rule \(s = s_N = \sqrt{\ln N}\), where \(N\) is the number of anchor points that instantiate kernels.
\Cref{eq:sigma_kk_mod} then depends on the data only through the dimensionless ratio \(d_{ik}^* / \tilde{d}\), scaled by the squared local length unit \(\tilde{d}^2\): densely supported anchor points receive the maximum kernel width \(s \tilde{d}\), i.e., \(s\) median neighbor gaps, wide enough to bridge the gaps between neighboring anchor points but no wider, while isolated anchor points receive exponentially narrower kernels, limiting their influence.
The calibration is defined with respect to the global nearest neighbor: per-dimension distances measure the spacing of one-dimensional projections, which shrinks with the inverse number of anchor points and vanishes entirely for quantized parameters, and therefore provides no meaningful local length scale.
At least two anchor points are required for the calibration to be defined.
The schedule \(s_N\), under which \cref{thm:AsympODDSimCalibrated} establishes two-sided convergence, is used for all calibrated experiments unless stated otherwise.
\begin{proposition}[Scale Invariance of the Calibrated ODD]\label{prop:ScaleInvariance}
    Let \(T(\bm{x}) = \bm{A} \bm{x} + \bm{b}\) with \(\bm{A} = \diag(a_1, \dots, a_n)\), \(a_k > 0\), be an affine re-expression of the parameters.
    The affinity function derived with \cref{eq:sigma_kk_mod,eq:sigma_calibrated} satisfies \(\alpha'(T(\bm{x})) = \alpha(\bm{x})\) for all \(\bm{x}\) whenever \(T\) is a similarity transformation, i.e., \(a_1 = \cdots = a_n\).
    Moreover, under a per-dimension affine normalization with equivariant defining bounds, the full derivation pipeline is invariant under arbitrary per-dimension affine re-expressions of the raw data.
\end{proposition}
The proof is given in \cref{sec:ScaleInvarianceProof}.
In particular, under such a normalization, the derived ODD does not depend on the choice of units---a property the fixed parameterization \(\kappa = \eta = 1\) does not possess.

\subsection{Handling Out-of-Distribution Data}
Often, the data used to define the data-driven ODD include not only ID samples as anchor points but also OOD samples explicitly outside the ODD\@.
Here, it is important to ensure that these points are not later classified as part of the ODD\@.
However, most kernels, especially the RBF kernel, cannot be zero anywhere; instead, a maximum affinity score can be defined that, at the locations of OOD samples, cannot be violated.
Inspired by the Development Assurance Levels~\cite{SAE-ARP4754B}, depending on the classification of failure conditions of the AI-based systems, different maximum values can be defined.\footnote{%
    Here, \enquote{fully automated} means execution given documented assurance inputs.
    The acquisition-resolution metadata used for de-duplication are likewise documented assurance inputs obtained from the recording-system specification, not parameters tuned against evaluation performance.
    The parameters \(\kappa{}\), \(\eta{}\), and \(\lambda{}\) are calibrated from data through \cref{eq:sigma_calibrated} using fixed \(\gamma{}\) and \(\lambda_{\mathrm{rel}}\) and the scheduling rule \(s_N = \sqrt{\ln N}\).
    The membership threshold \(\zeta{}\) is conformally calibrated for a documented false-exclusion budget \(\varepsilon{}\) (cf.\ \cref{subsec:ConformalThreshold}), while \(c \in (0, 1)\) is a fixed algorithmic step-size parameter affecting adjustment granularity and convergence rate.
    The OOD affinity bound \(\xi{}\) remains a documented assurance decision because the Development Assurance Levels prescribe no numerical affinity bound for a data-driven, kernel-based ODD\@.
}
After calculating the kernel parameters purely on ID samples, the global affinity function \(\alpha(\bm{x})\) has to be sampled at all OOD sample locations.
If the affinity is above the defined threshold, the most influential kernels have to be tuned such that \(\alpha(\bm{x})\) stays below the threshold.
This adjustment constrains the affinity only at observed OOD locations; unsafe regions absent from the OOD set are not guaranteed to be excluded, so the constraint complements rather than replaces expert-specified exclusions (cf.\ \cref{subsec:Limitations}).

\subsection{Creation of the Kernel-Based ODD}
All previously mentioned steps are combined in \cref{alg:autoSAFE}, which defines how to derive the ODD from data.
First, the available records are collected and split into ID and OOD classes, and a held-out calibration set is reserved from the ID records.
The remaining ID records are then de-duplicated using the documented acquisition resolutions, before anchor subsampling and kernel calibration.
The held-out calibration records do not instantiate kernels and retain their original sampling frequencies, while every OOD sample remains in the consistency check.
Next, a kernel is chosen, and its parameters are calibrated from the retained ID anchors as described in \cref{subsec:KernelParameters}.
After the OOD consistency adjustment, \(\zeta{}\) is calibrated from the held-out ID set for the documented false-exclusion budget \(\varepsilon{}\), as described in \cref{subsec:ConformalThreshold}.
For \(L\) input ID records, deterministic cell construction and representative selection require at most \(O(Ln \log L)\) preprocessing time and \(O(Ln)\) space.
The naïve algorithm (cf.\ \cref{alg:autoSAFE}) runs in \(O(N^2 n + N_{\mathrm{adj}}MNn + N_{\mathrm{cal}}Nn)\) time and requires \(O(Nn)\) space with \(N\) retained anchors, \(M\) OOD samples, \(N_{\mathrm{adj}}\) OOD-correction steps, and \(N_{\mathrm{cal}}\) calibration records.
Each subsequent query evaluates in \(O(Nn)\) time.
The individual terms and optional theoretical accelerations are discussed in \cref{sec:ComputationalScalability}.

\section{Validation}\label{sec:Validation}
As part of this work, the autoSAFE tool was developed and open-sourced\footnote{GitHub repository: \url{https://github.com/DLR-KI/autosafe}} to validate the proposed method for deriving the ODD \(\mathcal{O}\) from data.
Implementation details are given in \cref{sec:ImplDetails}; hardware, runtimes, normalization, and default parameterizations are documented in \cref{subsec:ExperimentalSetup}, with experiment-specific variations reported alongside their results.

\subsection{Synthetic Benchmarks}\label{subsec:MCSampling}
The synthetic validation covers analytically specified half-space, annular, two-blob, banana-shaped, five-dimensional constrained, and rotated-band ODDs, including Monte Carlo configurations from two to twelve dimensions.
The experiments examine anchor-count and parameter sensitivity, boundary halos, OOD adjustment, membership-threshold calibration, sampling density and duplicates, anchor-subset and permutation stability, covariance structure, and truncated-affinity scalability.
Complete setups, hyperparameters, and results are reported in \cref{sec:AdditionalValidation,sec:JustDiagSigma,sec:ComputationalScalability}.
The two-dimensional suite compares autoSAFE with kernel density estimation (KDE)~\cite{Devroye1980}, a Gaussian mixture model (GMM) selected using the Bayesian information criterion (BIC)~\cite{McLachlan2000, Schwarz1978}, \(k\)-nearest-neighbor (\(k\)-NN) distance, a one-class support vector machine (OC-SVM)~\cite{Schoelkopf2001}, Isolation Forest~\cite{Liu2008}, and the convex hull; at \(N = \num{1000}\), autoSAFE remains within \num{0.01548} area under the precision--recall curve (AUPR) of the strongest baseline and substantially outperforms the convex hull on the non-convex domains.
Across the remaining tests, calibrated bandwidths control exterior expansion as coverage grows, all 72 OOD-adjustment configurations satisfy the specified affinity bound, conformal false-exclusion rates track their targets, and permutation-induced affinity differences remain at the level of floating-point rounding; in the sparse five-dimensional case, the conformal threshold keeps its target false-exclusion rate, but the region at a fixed threshold strongly depends on \(s\).

\subsection{Real-World Use Case}\label{subsec:ValidationviaUseCase}
In aviation, one of the most important tasks of pilots during en route travel is avoiding collisions with other aircraft.
The next-generation system, Advanced Collision Avoidance System X (ACAS X), is currently under development and expected to contain an AI-based system~\cite{ED-256, ED-275, DO-385, DO-386}.
Preliminary implementations split the system into a vertical and a horizontal component, called the Vertical Collision Avoidance System (VCAS) and the Horizontal Collision Avoidance System (HCAS), respectively~\cite{Katz2017, Julian2019, ManzanasLopez2023}; see \cref{sec:ACASX} for details.
Based on parameter ranges required for safe operation and previous experiments from other works~\cite{Stefani2024b, Christensen2024a}, a dataset of all observed state vectors has been created for each component, with \num{622110} anchor points for VCAS and \num{385925} for HCAS.
Next, they were used to create the data-driven ODD \(\mathcal{O}\).
Once the data-driven ODD had been created, \num{e6} validation samples were created: one half drawn uniformly from the parameter ranges defined in \cref{tab:VCAS_ODD,tab:HCAS_ODD}, expanded by \qty{10}{\percent} per continuous dimension, and one half by perturbing randomly selected anchor points with isotropic Gaussian noise of per-dimension standard deviation \(3\tilde{d}/\sqrt{n}\), so that the expected displacement is comparable to the calibrated kernel scale.
The discrete values of \(s_\mathrm{adv}\) are not perturbed and only take their valid values.
As before, for each validation sample, the affinity value and its membership in the known ODD and in the anchor-derived reference regions of \cref{tab:AviationReferenceAgreement} were calculated.
Subsequently, the confusion matrices were computed for different affinity thresholds, and precision and recall vs.\ affinity threshold plots were generated.
The thresholds are spaced adaptively toward both decision extremes and are evaluated directly through the log-survival rule of \cref{eq:LogThreshold}; \cref{subsec:ThresholdGrid} gives the exact grid.
The resulting plots for VCAS and HCAS are shown in \cref{fig:CAS-PR}, together with the split-conformal threshold \(\hat{\zeta}_{0.1}\) for a false-exclusion budget of \qty{10}{\percent} (cf.\ \cref{subsec:ConformalThreshold}), which is \num{0.9923} and \num{0.9861}, for VCAS and HCAS, respectively.
Precision remains above \num{0.99} at every threshold, whereas accepting every validation sample would only reach the shares of samples inside the known ODD, \num{0.789} for VCAS and \num{0.738} for HCAS.
Between the smallest positive threshold and \(\hat{\zeta}_{0.1}\), recall decreases from \num{0.633} to \num{0.592} and from \num{0.655} to \num{0.461}, for VCAS and HCAS, respectively.
Recall is lower than in the Monte Carlo example because the recorded states do not visit large parts of the known ODD, so many validation samples inside it lie far from every anchor point.
Because the known VCAS ODD is available here, the curves are evaluated directly against that reference.
In applications without independently specified ODD membership information, affinity thresholds instead require held-out validation data with the relevant labels.
A separate VCAS OOD experiment selected \num{20000} ID anchors uniformly at random without replacement from the \num{559918} records outside an excluded region; here, anchor subsampling means that only the selected ID records instantiate kernels, while all \num{62192} records in the excluded region remain OOD constraints.
The experiment used \num{200000} evaluation samples and terminated after \num{4389} adjustment steps, modifying \num{235} kernels and reducing the maximum observed OOD affinity to \num{0.00997} for \(\xi = \num{0.01}\).
This result is explicitly scoped to the subsampled VCAS construction; complementary full-data HCAS reference-agreement, seven-reference aviation, and synthetic and real-data anchor-subset results are reported in \cref{tab:AviationReferenceAgreement,fig:SubsetStability,tab:AviationSubsetStability}.

\section{Discussion}\label{sec:Discussion}
The presented results demonstrate that the proposed kernel-based approach can effectively approximate an underlying ODD from data alone, given labeled ID samples and the calibration procedure of \cref{subsec:KernelParameters}.
This capability is crucial for legacy systems or complex environments where an analytic description of the ODD is unavailable or incomplete.
A key finding is that agreement with a reference ODD can be quantified on held-out data through the affinity-threshold sweep.

The convex hull is generally inadequate as the final ODD representation.
Real-world operational parameters often form non-convex, manifold-like structures with \emph{holes} or concavities (e.g., specific combinations of altitude and speed that are physically impossible or unsafe).
A convex hull heedlessly encloses these void regions, potentially deeming the system safe in conditions where no data exist, and safety cannot be verified.
In contrast, the region represented by the proposed kernel-based method follows the anchor points and largely excludes non-convex void regions, up to a boundary layer whose width depends on the local anchor spacing.
On held-out points from the excluded VCAS region, the convex hull has a false-positive rate of \num{0.547}, compared with \num{0.042} for autoSAFE without OOD samples and \num{0.0055} after adding the region's points as OOD constraints (cf.\ \cref{tab:HeldOutHole}).
At the fixed threshold \(\zeta = 0.5\), the intersection over union (IoU) on the two-blob and banana ODDs is \num{0.855} and \num{0.775} for autoSAFE, compared with \num{0.488} and \num{0.338} for the convex hull; on the annulus, autoSAFE's IoU of \num{0.892} versus \num{0.898} for the hull is accompanied by a smaller mean false-positive boundary distance of \num{0.144} versus \num{0.358}, distinguishing a thin exterior halo from errors inside the excluded hole (cf.\ \cref{tab:BaselineFixedThreshold}).
With the calibrated parameters of \cref{eq:sigma_calibrated}, the affinity boundary layer scales with the local anchor spacing, so the representation is conservative in sparsely covered regions and tightens around the data as coverage increases (cf.\ \cref{subsec:Limitations,thm:AsympODDSimCalibrated}).

The OOD consistency adjustment was exercised on synthetic half-space and annular ODDs with \(N = \num{1000}\) anchors and \(M \in \{10, 100, 1000\}\) uniformly sampled OOD points.
To isolate the OOD-adjustment behavior, the 72 tested configurations use the fixed evaluation threshold \(\zeta = 0.6\) and cover \(\xi \in \{0.01, 0.1, 0.3\}\) and \(c \in \{0.5, 0.8, 0.9, 0.99\}\).
In every configuration, the adjustment terminated with all observed OOD affinities below \(\xi{}\), after \numrange{0}{32551} iterations; across the grid, a single kernel was rescaled at most \num{553} times.
Collateral effects were evaluated on separately sampled held-out points from the analytically specified in-distribution region, not on the anchors.
At most \qty{5.0}{\percent} of the held-out points that passed \(\zeta{}\) before adjustment fell below it afterward.
Across the same evaluation, recall decreased by at most \num{0.047}, while precision increased by up to \num{0.090} as the over-inflated boundary tightened.

A held-out conformal calibration set determines \(\zeta{}\) from data: across the tested error budgets, the empirical false-exclusion rate closely follows the target.
For a documented false-exclusion budget \(\varepsilon{}\), \(\zeta{}\) is selected from a held-out ID calibration set so that, under exchangeability, a future ID sample falls below the threshold with probability at most \(\varepsilon{}\) (cf.\ \cref{subsec:ConformalThreshold}).

Furthermore, the continuous nature of the affinity function \(\alpha(\bm{x})\) introduces a built-in mechanism for outlier detection and runtime monitoring.
Unlike rigid geometric boundaries that provide a binary inside/outside classification, the affinity score offers a measure of proximity to known safe conditions.
As the system state drifts away from the anchor points, \(\alpha(\bm{x})\) decays smoothly.
This property enables the definition of graded warning zones, allowing the system to trigger degradation modes or alert operators before the hard safety boundary is crossed.
This \emph{soft} boundary is a distinct advantage of the kernel-based representation over traditional Boolean logic ODD definitions.

\subsection{Limitations}\label{subsec:Limitations}
The presented kernel-based representation does have some limitations.
First, the affinity depends on the selected kernel parameters; for the RBF kernel, these are \(\kappa{}\), \(\eta{}\), and \(\lambda{}\).
The calibration in \cref{eq:sigma_calibrated} removes dependence on coordinate scale and reduces the remaining choice to the dimensionless constants \(\gamma{}\) and \(\lambda_{\mathrm{rel}}\).
For the tested two-dimensional ODDs with \(N = \num{1000}\), the calibrated parameter grid produces an absolute AUPR range of at most \num{0.0252}.
On the linear ODD, fixed pairs of \(\kappa{}\) and \(\eta{}\) on normalized data rank the validation points similarly well, with AUPR values of \numrange{0.966}{0.995}, but their level sets at \(\zeta = 0.5\) reach an IoU of only \numrange{0.189}{0.294}, compared with \numrange{0.859}{0.902} for \(s_N\), because fixed values implicitly select a unit of length (cf.\ \cref{fig:ParameterSensitivity}).
The sparse five-dimensional constrained ODD is more sensitive: its AUPR stays within \numrange{0.825}{0.909} across the calibrated grid, but at \(\zeta = 0.5\) the IoU reaches at most \num{0.029} for \(s_N\), compared with up to \num{0.591} for \(s = 1\).
This highlights both the need for the conformal threshold calibration of \cref{subsec:ConformalThreshold} and the potential value of a dimension-dependent scheduling rule as discussed in \cref{subsec:ParameterSensitivity}.
Second, the affinity-based representation has two density regimes.
In sparsely covered regions, it can under-approximate the intended ODD, whereas near densely covered boundaries, fixed kernel parameters can over-approximate it through accumulated kernel tails.
\Cref{prop:AsympODDSim} proves eventual interior coverage and bounds the represented region by an outer halo, but it does not establish two-sided set convergence at fixed parameters.
The calibrated parameterization instead guarantees that the global containment radius shrinks to zero and that the symmetric-difference volume converges to zero under the assumptions of \cref{thm:AsympODDSimCalibrated}.
The measured fixed and calibrated regimes are shown in \cref{fig:HaloAnchorCount}.
Third, the derived ODD reflects where operational data are dense, not which operating conditions are safe in a causal or system-level sense.
Operating conditions that are unsafe yet present in the recorded data remain inside the ODD, and valid conditions absent from the data are excluded; establishing the safety of the intended functionality within the ODD remains a separate assurance activity (cf.\ ISO~21448~\cite{ISO21448} and UL~4600~\cite{UL4600}).
Furthermore, the validation in \cref{sec:Validation} draws its samples from the same generating processes as the anchor data; robustness to distribution shift between development and operation, and to mislabeled ID or OOD samples, has not been evaluated.
The data-driven ODD should therefore be read as one reproducible evidence artifact within a broader safety case (cf.\ \cref{subsec:CertificationImpact}), not as certification-level safety evidence on its own.
Resolution-cell de-duplication depends on reliable acquisition metadata.
A cell width larger than the actual recording resolution can merge genuinely distinct operational states, whereas a smaller width can preserve information-free multiplicity.
The selected resolutions and resulting merge counts must therefore be documented as part of the data-assurance record.
Nevertheless, this approach is not meant to deprecate outright the development of ODDs based on expert knowledge.
It is meant to support current best practices in ODD development and to bolster development in areas with little to no expert knowledge.

\subsection{Certification Impact}\label{subsec:CertificationImpact}
The proposed method provides a reproducible evidence artifact for the overall assurance and certification of AI-based systems.
A well-defined ODD serves as the basis for the development of any AI-based system, whether in automotive~\cite{BSI2020, ISO34503} or aviation~\cite{EUASA2024}.
Given the deterministic and explainable approach provided in this work, the derived ODD delineates the operational limits within which the safe deployment of an AI-based system can be argued.
This, for example, feeds into EASA's learning assurance and its requirement for a continuous safety assessment.
Moreover, the kernel-based representation can be used as a monitoring tool to detect whether the system remains within the defined boundaries.
Thus, the proposed kernel-based approach supplements the proactive development of safety-by-design AI-based systems and the continuous monitoring of deployed systems, functioning as an early-detection mechanism before the system leaves its operational boundaries.
\Cref{sec:EASAObjectives} details selected objectives of the EASA concept paper~\cite{EUASA2024} that the presented framework supports.

\section{Conclusion}\label{sec:Conclusion}
This work showed that, although a fully truthful data-driven reconstruction of an ODD is not always possible, the proposed approach can sufficiently approximate the underlying ODD\@.
This work also presents a formal, data-centric definition of ODD similarity, necessary for future data-driven ODD development.
Importantly, the presented method yields an ODD representation that is deterministic, permutation-stable, and conservative in sparsely covered regions (cf.\ \cref{rem:TwoRegime}), addressing key safety-by-design requirements on the path toward certification.
By using a kernel-based representation, the framework provides a mathematically rigorous means to define operational boundaries that are interpretable and tractable.
The novel, safety-by-design, deterministic kernel-based affinity representation of ODDs is derived fully automatically, given documented assurance inputs, via an algorithm that is permutation-stable, bounded by construction, and, under affine-equivariant per-dimension normalization, invariant to the choice of units.
Unlike prior data-centric approaches~\cite{Kaakai2023}, the proposed method makes no assumption about the parametric form of the ODD boundary, is permutation-stable with respect to sample ordering, and produces an affinity function that is bounded by construction and amenable to formal safety argumentation.
Future work will focus on optimizing the kernel hyperparameter selection process.
While this work focused on diagonal kernel parameter matrices to enhance robustness and interpretability, selectively extending the framework to model cross-dimensional dependencies could further improve expressiveness in tightly coupled systems.
Such extensions would require careful constraints to preserve determinism and certifiability.
Moreover, integrating temporal information through time-dependent kernels could enable the representation of evolving or context-dependent ODDs.
Finally, future work should explore tighter integration of kernel-based ODDs into certification workflows.
This includes linking affinity thresholds to formal safety requirements, leveraging the representation for runtime assurance monitors, and studying how data-driven ODD updates can be performed in a controlled and certifiable manner.
Such developments would further strengthen the role of data-driven ODDs in deploying safety-critical AI-based systems.

\clearpage{}

\begin{ack}
    This work received no external funding.
    The authors declare no competing interests.
\end{ack}

\bibliographystyle{IEEEtranN}
\bibliography{literature-bibtex}

\appendix{}
\crefalias{section}{appendix}
\crefalias{subsection}{subappendix}
\crefalias{subsubsection}{subsubappendix}



\section{Additional Validation Results}\label{sec:AdditionalValidation}

\subsection{Experimental Setup}\label{subsec:ExperimentalSetup}
All experiments were conducted on a system with an Intel Core i9-13900K, an NVIDIA RTX 4090, and \qty{64}{\giga\byte} of RAM using Python 3.14.3.
The original Monte Carlo experiments completed in under \qty{5}{\minute}, the additional validation experiments individually required between \qty{9}{\second} and \qty{15}{\minute}, and the aviation experiments approximately \qtyrange{11}{14}{\hour} each.
The original Monte Carlo experiments used the manual parameterization \(\kappa = 1\), \(\eta = 1\), and \(\lambda_{\mathrm{rel}} = \exp(-10)\) on the raw parameter space.
Unless stated otherwise in the corresponding experiment description, the calibrated synthetic experiments used per-dimension normalization to \([-1, 1]\), \(\gamma = 1\), the scheduling rule \(s = s_N = \sqrt{\ln N}\), and \(\lambda_{\mathrm{rel}} = \exp(-10)\).
Threshold-free metrics such as the AUPR rank the validation points by the log-survival score \(-S(\bm{x})\) of \cref{eq:LogSurvival}, which orders them exactly as \(\alpha{}\) does but stays distinguishable where \(\alpha{}\) rounds to one in finite precision.
For the aviation experiments, normalization was based on the known ODD definitions derived from prior research~\cite{Julian2019} and summarized in \cref{tab:HCAS_ODD,tab:VCAS_ODD}.

\subsection{Monte Carlo Validation}\label{subsec:MonteCarloValidation}
Monte Carlo methods (MCM)~\cite{Metropolis1949} have proven useful for numerical problem solving.
Here, MCM is used to generate synthetic data for analytically specified ODDs by sampling anchor points uniformly within each ODD and validation points uniformly from a larger enclosing domain, thereby enabling affinity-threshold membership to be compared directly with known ODD membership.
The experiments cover ODD structures from two to twelve dimensions.
The two-dimensional anchor-count sweep described below additionally isolates the effect of data coverage from the sparsest tested case to dense sampling.

A representative two-dimensional case is depicted in \cref{fig:2D-MCM-ODD}, where the blue dots are the anchor points, the red dots are the validation samples, the black rectangle is the polytope \(X\), and the black line is the boundary specified by \(R_1\).
For this setup, the data-driven ODD \(\mathcal{O}\) was derived according to \cref{alg:autoSAFE}, although with manual parameterization, across anchor counts ranging from the sparsest tested configuration to dense sampling.
More precisely, \(X\) was defined as \([-5, 5] \times [-5, 5]\), \(R^\mathcal{O}_1 = \{\bm{x} \in X \mid x_2 \geq x_1 - 3\}{}\), and the tested anchor counts were \(N \in \{3, 5, 10, 25, 30, 50, 100, 300, \num{1000}, \num{3000}, \num{10000}\}\).
For each anchor count, five independently seeded runs were performed, each using a newly sampled anchor set and \num{100000} validation points.
The large validation set was chosen to provide coverage inside and outside the analytically specified ODD, including regions close to the anchors.
The validation points were drawn uniformly from \([-10, 10] \times [-10, 10]\), whose side lengths are twice those of \(X\).

\begin{figure}[htb]
    \begin{subfigure}[t]{0.475\linewidth}
        \centering
        \resizebox{\linewidth}{!}{%
            \begin{tikzpicture}
    \begin{axis}[
        autosafe plot,
        axis lines=center,
        axis line style={-Latex[round]},
        xlabel={\(x_1\)},
        ylabel={\(x_2\)},
        x label style={
                at={(axis description cs:0.5,0)},
                anchor=north,
            },
        y label style={
                at={(axis description cs:0,0.5)},
                rotate=90,
                anchor=south,
            },
        xtick=\empty,
        ytick=\empty,
        xmin=0,
        xmax=1,
        ymin=0,
        ymax=1,
        ]

        \draw [draw=black] (0.25,0.25) rectangle (0.75,0.75);
        \addplot [color = black, name path=A] {x - 3/10};
        \addplot [draw = none, name path=B] {x - 6/10};
        \addplot [
            pattern color=black!70!white,
            pattern=north west lines,
        ] fill between [
            of=A and B,
        ];

        \node at (0.31, 0.62) [circle, fill=blue, inner sep=1.5pt]{};
        \node at (0.39, 0.29) [circle, fill=blue, inner sep=1.5pt]{};
        \node at (0.52, 0.43) [circle, fill=blue, inner sep=1.5pt]{};
        \node at (0.61, 0.68) [circle, fill=blue, inner sep=1.5pt]{};
        \node at (0.7, 0.5) [circle, fill=blue, inner sep=1.5pt]{};
        %
        \node at (0.42, 0.61) [circle, fill=blue, inner sep=1.5pt]{};
        \node at (0.29, 0.33) [circle, fill=blue, inner sep=1.5pt]{};
        \node at (0.44, 0.71) [circle, fill=blue, inner sep=1.5pt]{};
        \node at (0.29, 0.48) [circle, fill=blue, inner sep=1.5pt]{};

        \node at (0.11, 0.075) [circle, fill=red, inner sep=1.5pt]{};
        \node at (0.15, 0.45) [circle, fill=red, inner sep=1.5pt]{};
        \node at (0.175, 0.8) [circle, fill=red, inner sep=1.5pt]{};
        \node at (0.27, 0.35) [circle, fill=red, inner sep=1.5pt]{};
        \node at (0.375, 0.11) [circle, fill=red, inner sep=1.5pt]{};
        \node at (0.41, 0.55) [circle, fill=red, inner sep=1.5pt]{};
        \node at (0.43, 0.9) [circle, fill=red, inner sep=1.5pt]{};
        \node at (0.5, 0.85) [circle, fill=red, inner sep=1.5pt]{};
        \node at (0.6, 0.06) [circle, fill=red, inner sep=1.5pt]{};
        \node at (0.6, 0.38) [circle, fill=red, inner sep=1.5pt]{};
        \node at (0.6, 0.6) [circle, fill=red, inner sep=1.5pt]{};
        \node at (0.7, 0.41) [circle, fill=red, inner sep=1.5pt]{};
        \node at (0.75, 0.15) [circle, fill=red, inner sep=1.5pt]{};
        \node at (0.81, 0.24) [circle, fill=red, inner sep=1.5pt]{};
        \node at (0.85, 0.405) [circle, fill=red, inner sep=1.5pt]{};
        \node at (0.93, 0.81) [circle, fill=red, inner sep=1.5pt]{};
        \node at (0.65, 0.275) [circle, fill=red, inner sep=1.5pt]{};

    \end{axis}
\end{tikzpicture}%
        }%
        \caption{Setup for comparing the data-driven ODD \(\mathcal{O}\) to the underlying original ODD\@. The black rectangle represents the polytope \(X\) and the black line depicts the inequality \(R_1\).}\label{fig:2D-MCM-ODD}
    \end{subfigure}
    \hfill
    \begin{subfigure}[t]{0.475\linewidth}
        \centering
        \resizebox{\linewidth}{!}{%
            \begin{tikzpicture}
    \begin{axis}[
        autosafe plot,
        axis lines=center,
        axis line style={-Latex[round]},
        xlabel={\(x_1\)},
        ylabel={\(x_2\)},
        x label style={
                at={(axis description cs:0.5,0)},
                anchor=north,
            },
        y label style={
                at={(axis description cs:0,0.5)},
                rotate=90,
                anchor=south,
            },
        xtick=\empty,
        ytick=\empty,
        xmin=0,
        xmax=1,
        ymin=0,
        ymax=1,
        ]

        \draw [draw=black!35!white] (0.25,0.25) rectangle (0.75,0.75);
        \addplot [color = black!35!white, name path=A] {x - 3/10};
        \addplot [draw = none, name path=B] {x - 6/10};
        \addplot [
            pattern color=black!35!white,
            pattern=north west lines,
        ] fill between [
            of=A and B,
        ];

        \node (AC1) at (0.31, 0.62) [circle, fill=blue, inner sep=1.5pt]{};
        \node (AC2) at (0.39, 0.29) [circle, fill=blue, inner sep=1.5pt]{};
        \node (AC3) at (0.52, 0.43) [circle, fill=blue, inner sep=1.5pt]{};
        \node (AC4) at (0.61, 0.68) [circle, fill=blue, inner sep=1.5pt]{};
        \node (AC5) at (0.7, 0.5) [circle, fill=blue, inner sep=1.5pt]{};
        \node (AC6) at (0.29, 0.33) [circle, fill=blue, inner sep=1.5pt]{};
        \node (AC7) at (0.29, 0.48) [circle, fill=blue, inner sep=1.5pt]{};
        \node (AC8) at (0.42, 0.61) [circle, fill=blue, inner sep=1.5pt]{};
        \node (AC9) at (0.44, 0.71) [circle, fill=blue, inner sep=1.5pt]{};
         \draw [draw=black] (AC1) -- (AC7);
        \draw [draw=black] (AC1) -- (AC9);
        \draw [draw=black] (AC2) -- (AC5);
        \draw [draw=black] (AC2) -- (AC6);
        \draw [draw=black] (AC4) -- (AC5);
        \draw [draw=black] (AC4) -- (AC9);
        \draw [draw=black] (AC6) -- (AC7);

        \node at (0.11, 0.075) [circle, fill=red, inner sep=1.5pt]{};
        \node at (0.15, 0.45) [circle, fill=red, inner sep=1.5pt]{};
        \node at (0.175, 0.8) [circle, fill=red, inner sep=1.5pt]{};
        \node at (0.27, 0.35) [circle, fill=red, inner sep=1.5pt]{};
        \node at (0.375, 0.11) [circle, fill=red, inner sep=1.5pt]{};
        \node at (0.41, 0.55) [circle, fill=red, inner sep=1.5pt]{};
        \node at (0.43, 0.9) [circle, fill=red, inner sep=1.5pt]{};
        \node at (0.5, 0.85) [circle, fill=red, inner sep=1.5pt]{};
        \node at (0.6, 0.06) [circle, fill=red, inner sep=1.5pt]{};
        \node at (0.6, 0.38) [circle, fill=red, inner sep=1.5pt]{};
        \node at (0.6, 0.6) [circle, fill=red, inner sep=1.5pt]{};
        \node at (0.7, 0.41) [circle, fill=red, inner sep=1.5pt]{};
        \node at (0.75, 0.15) [circle, fill=red, inner sep=1.5pt]{};
        \node at (0.81, 0.24) [circle, fill=red, inner sep=1.5pt]{};
        \node at (0.85, 0.405) [circle, fill=red, inner sep=1.5pt]{};
        \node at (0.93, 0.81) [circle, fill=red, inner sep=1.5pt]{};
        \node at (0.65, 0.275) [circle, fill=red, inner sep=1.5pt]{};
    \end{axis}
\end{tikzpicture}%
        }%
        \caption{Setup for comparing the data-driven ODD \(\mathcal{O}\) to the convex hull of all anchor points.}\label{fig:2D-MCM-Hull}
    \end{subfigure}
    \caption{Setup for the two-dimensional example of the Monte Carlo method. The blue dots are the anchor points from which the data-driven ODD \(\mathcal{O}\) is derived. The red dots are the validation samples to compare the data-driven ODD \(\mathcal{O}\) to the underlying original ODD\@.}\label{fig:2D-MCM}
\end{figure}
Next, for all validation points and all anchor point configurations, the affinity and whether each point lies within the original ODD were calculated.
Based on the data, the confusion matrices were computed for different affinity thresholds.
The evaluation uses an adaptive log-space threshold grid whose exact construction is given in \cref{subsec:ThresholdGrid}.
Here, the affinity threshold determines whether a validation sample point is classified as inside the data-driven ODD\@.
Validation samples with affinity values at or above the affinity threshold \(\zeta{}\) are considered within the data-driven ODD; those below it are not. 
Given the results, precision and recall vs.\ affinity threshold plots were created (cf.\ \cref{fig:MCM-PerN}).
With these fixed kernel parameters, increasing the number of anchor points decreases precision while increasing recall, consistent with the outer halo of \cref{prop:AsympODDSim}; the calibrated parameters avoid this expansion (cf.\ \cref{fig:HaloAnchorCount}).

\begin{figure}[htb]
    \begin{subfigure}[t]{0.475\linewidth}
        \centering
        \resizebox{\linewidth}{!}{%
            \begin{tikzpicture}
    \begin{axis}[
        autosafe plot,
        xlabel={Affinity Threshold \(\zeta{}\)},
        ylabel={Precision},
        xtick={0,0.2,0.4,0.6,0.8,1.0},
        ytick={0,0.5,1.0},
        xmin=0,
        xmax=1,
        ymin=0,
        ymax=1,
        filter discard warning=false,
        legend style={
            at={(0.5,-0.28)},
            anchor=north,
            legend columns=4,
            font=\scriptsize,
            draw=none,
            fill=none,
            /tikz/every even column/.append style={column sep=4pt},
        },
    ]
        \addplot[color=DLRDarkerBlue,mark=none,solid,discard if not={n}{3}]
            table[x=zeta,y=precision_mean]{graphics/data/benchmark/mcm_per_n.dat};
        \addlegendentry{\(3\)}
        \addplot[color=DLRDarkerBlue,mark=none,densely dashed,discard if not={n}{5}]
            table[x=zeta,y=precision_mean]{graphics/data/benchmark/mcm_per_n.dat};
        \addlegendentry{\(5\)}
        \addplot[color=DLRDarkerBlue,mark=none,densely dotted,discard if not={n}{10}]
            table[x=zeta,y=precision_mean]{graphics/data/benchmark/mcm_per_n.dat};
        \addlegendentry{\(10\)}
        \addplot[color=DLRDarkBlue,mark=none,solid,discard if not={n}{25}]
            table[x=zeta,y=precision_mean]{graphics/data/benchmark/mcm_per_n.dat};
        \addlegendentry{\(25\)}
        \addplot[color=DLRDarkBlue,mark=none,densely dashed,discard if not={n}{30}]
            table[x=zeta,y=precision_mean]{graphics/data/benchmark/mcm_per_n.dat};
        \addlegendentry{\(30\)}
        \addplot[color=DLRDarkBlue,mark=none,densely dotted,discard if not={n}{50}]
            table[x=zeta,y=precision_mean]{graphics/data/benchmark/mcm_per_n.dat};
        \addlegendentry{\(50\)}
        \addplot[color=DLRDarkerGreen,mark=none,solid,discard if not={n}{100}]
            table[x=zeta,y=precision_mean]{graphics/data/benchmark/mcm_per_n.dat};
        \addlegendentry{\(100\)}
        \addplot[color=DLRDarkerGreen,mark=none,densely dashed,discard if not={n}{300}]
            table[x=zeta,y=precision_mean]{graphics/data/benchmark/mcm_per_n.dat};
        \addlegendentry{\(300\)}
        \addplot[color=DLRDarkerGreen,mark=none,densely dotted,discard if not={n}{1000}]
            table[x=zeta,y=precision_mean]{graphics/data/benchmark/mcm_per_n.dat};
        \addlegendentry{\(1000\)}
        \addplot[color=DLRDarkerYellow,mark=none,densely dashed,discard if not={n}{3000}]
            table[x=zeta,y=precision_mean]{graphics/data/benchmark/mcm_per_n.dat};
        \addlegendentry{\(3000\)}
        \addplot[color=DLRDarkerYellow,mark=none,solid,discard if not={n}{10000}]
            table[x=zeta,y=precision_mean]{graphics/data/benchmark/mcm_per_n.dat};
        \addlegendentry{\(10000\)}
    \end{axis}
    \autosafepairedplotbounds{48pt}{11pt}{75pt}{8pt}
\end{tikzpicture}%
        }%
        \caption{Mean precision against known ODD membership for each tested anchor count.}\label{fig:MCM-PerN-Precision}
    \end{subfigure}
    \hfill
    \begin{subfigure}[t]{0.475\linewidth}
        \centering
        \resizebox{\linewidth}{!}{%
            \begin{tikzpicture}
    \begin{axis}[
        autosafe plot,
        xlabel={Affinity Threshold \(\zeta{}\)},
        ylabel={Recall},
        xtick={0,0.2,0.4,0.6,0.8,1.0},
        ytick={0,0.5,1.0},
        xmin=0,
        xmax=1,
        ymin=0,
        ymax=1,
        filter discard warning=false,
        legend style={
            at={(0.5,-0.28)},
            anchor=north,
            legend columns=4,
            font=\scriptsize,
            draw=none,
            fill=none,
            /tikz/every even column/.append style={column sep=4pt},
        },
    ]
        \addplot[color=DLRDarkerBlue,mark=none,solid,discard if not={n}{3}]
            table[x=zeta,y=recall_mean]{graphics/data/benchmark/mcm_per_n.dat};
        \addlegendentry{\(3\)}
        \addplot[color=DLRDarkerBlue,mark=none,densely dashed,discard if not={n}{5}]
            table[x=zeta,y=recall_mean]{graphics/data/benchmark/mcm_per_n.dat};
        \addlegendentry{\(5\)}
        \addplot[color=DLRDarkerBlue,mark=none,densely dotted,discard if not={n}{10}]
            table[x=zeta,y=recall_mean]{graphics/data/benchmark/mcm_per_n.dat};
        \addlegendentry{\(10\)}
        \addplot[color=DLRDarkBlue,mark=none,solid,discard if not={n}{25}]
            table[x=zeta,y=recall_mean]{graphics/data/benchmark/mcm_per_n.dat};
        \addlegendentry{\(25\)}
        \addplot[color=DLRDarkBlue,mark=none,densely dashed,discard if not={n}{30}]
            table[x=zeta,y=recall_mean]{graphics/data/benchmark/mcm_per_n.dat};
        \addlegendentry{\(30\)}
        \addplot[color=DLRDarkBlue,mark=none,densely dotted,discard if not={n}{50}]
            table[x=zeta,y=recall_mean]{graphics/data/benchmark/mcm_per_n.dat};
        \addlegendentry{\(50\)}
        \addplot[color=DLRDarkerGreen,mark=none,solid,discard if not={n}{100}]
            table[x=zeta,y=recall_mean]{graphics/data/benchmark/mcm_per_n.dat};
        \addlegendentry{\(100\)}
        \addplot[color=DLRDarkerGreen,mark=none,densely dashed,discard if not={n}{300}]
            table[x=zeta,y=recall_mean]{graphics/data/benchmark/mcm_per_n.dat};
        \addlegendentry{\(300\)}
        \addplot[color=DLRDarkerGreen,mark=none,densely dotted,discard if not={n}{1000}]
            table[x=zeta,y=recall_mean]{graphics/data/benchmark/mcm_per_n.dat};
        \addlegendentry{\(1000\)}
        \addplot[color=DLRDarkerYellow,mark=none,densely dashed,discard if not={n}{3000}]
            table[x=zeta,y=recall_mean]{graphics/data/benchmark/mcm_per_n.dat};
        \addlegendentry{\(3000\)}
        \addplot[color=DLRDarkerYellow,mark=none,solid,discard if not={n}{10000}]
            table[x=zeta,y=recall_mean]{graphics/data/benchmark/mcm_per_n.dat};
        \addlegendentry{\(10000\)}
    \end{axis}
    \autosafepairedplotbounds{48pt}{11pt}{75pt}{8pt}
\end{tikzpicture}%
        }%
        \caption{Mean recall against known ODD membership for each tested anchor count.}\label{fig:MCM-PerN-Recall}
    \end{subfigure}
    \caption{Precision and recall vs.\ affinity threshold for the Monte Carlo anchor-count sweep. Each curve is the mean over five seeds for one anchor count, exposing the sparse-data regime and the progression with increasing coverage.}\label{fig:MCM-PerN}
\end{figure}
Moreover, the sharp edges where the affinity is either zero or one follow from \cref{eq:Superposition}: the RBF kernel cannot be exactly zero, so no validation sample attains an affinity of exactly zero, while an affinity of exactly one would, in exact arithmetic, require a validation sample to land precisely on an anchor point.
In finite-precision arithmetic, however, samples deep inside the ODD can saturate to an affinity of exactly one; the reported results therefore use the log-space rule of \cref{eq:LogThreshold}.
Given the precision and recall vs.\ affinity threshold plots, one can derive an affinity threshold for the use case by balancing precision and recall within overall constraints, such as required safety levels.

The underlying ODD is known for this synthetic validation.
However, it will generally not be available in a real-world application.
Therefore, the data-driven ODD is also compared with the convex hull of all anchor points as a geometric reference.
This approach is illustrated in \cref{fig:2D-MCM-Hull}, where the original ODD structure is still drawn, albeit faintly.
The same steps as before were repeated, and corresponding precision and recall vs.\ affinity threshold curves were computed using convex-hull membership as the reference.
The per-\(N\) analysis compares the convex hull--referenced curves with the underlying ODD--referenced curves on a shared threshold grid.
For two membership curves \(q^{\mathrm{ODD}}\) and \(q^{\mathrm{ref}}\), with \(q\) denoting precision or recall, the curve agreement is \(R^2 = 1 - \sum_j (q_j^{\mathrm{ref}} - q_j^{\mathrm{ODD}})^2/\sum_j (q_j^{\mathrm{ODD}} - \bar q^{\mathrm{ODD}})^2\), where the underlying-ODD curve is the target.

The precision-curve \(R^2\) is negative for \(N \leq 50\), e.g., \num{-3.06} at \(N = 25\), becomes positive at \(N = 100\), and reaches \num{0.971} at \(N = \num{1000}\) and \num{0.99965} at \(N = \num{10000}\), where the recall-curve \(R^2\) reaches \num{1.000} (cf.\ \cref{fig:PerNValidation}).
In this setting, the convex hull is therefore a usable proxy reference only beyond approximately \num{100} anchors.
The corresponding area under the precision--recall curve (AUPR) reaches \(\num{0.99902(11)}\), whereas at \(N = 5\), \(25\), and \(50\) it is \(\num{0.734(56)}\), \(\num{0.903(20)}\), and \(\num{0.949(12)}\), respectively.
These results quantify both the sparse-data regime and the observed improvement toward the underlying ODD over the tested anchor-count range.
At \(N = \num{1000}\), autoSAFE is within \num{0.01548} AUPR of the strongest tested baseline throughout the two-dimensional suite and substantially exceeds the convex hull on the non-convex annulus and two-blob domains (cf.\ \cref{tab:BaselineAUPR}).

The boundary-halo experiment in \cref{fig:HaloAnchorCount} complements the threshold-sweep results by directly measuring exterior expansion at \(\zeta = 0.5\) as the anchor count increases.
The fixed-parameter halo grows from \num{1.027} to \num{3.426}, whereas the calibrated halo shrinks to \num{0.102}.
These measurements illustrate why calibration is needed to control exterior expansion under denser sampling.

\begin{figure}[htb]
    \begin{subfigure}[t]{0.475\linewidth}
        \centering
        \resizebox{\linewidth}{!}{%
            \begin{tikzpicture}
    \begin{axis}[
            autosafe plot,
            xmode=log,
            log basis x=10,
            xmin=30,
            xmax=30000,
            xlabel={Anchor count \(N\)},
            ylabel={Maximum exterior halo},
            ymin=0,
            ymax=4,
            legend style={at={(0.98,0.42)},anchor=east},
        ]
        \addplot[color=DLRDarkerYellow,mark=triangle*] table[x=n,y=fixed_halo]{graphics/data/benchmark/halo_wide.dat};
        \addlegendentry{Fixed}
        \addplot[color=DLRDarkerBlue,mark=*] table[x=n,y=calibrated_sN_halo]{graphics/data/benchmark/halo_wide.dat};
        \addlegendentry{Calibrated}
    \end{axis}
    \autosafepairedplotbounds{40pt}{2pt}{34pt}{7pt}
\end{tikzpicture}%
        }%
        \caption{Maximum distance from a point classified as in the ODD at \(\zeta = 0.5\) to the known ODD boundary as the number of anchor points increases.}\label{fig:HaloAnchorCount-Halo}
    \end{subfigure}
    \hfill
    \begin{subfigure}[t]{0.475\linewidth}
        \centering
        \resizebox{\linewidth}{!}{%
            \begin{tikzpicture}
    \begin{axis}[
        autosafe plot,
        xmode=log,
        log basis x=10,
        xmin=30,
        xmax=30000,
        xlabel={Anchor count \(N\)},
        ylabel={Precision at \(\zeta = 0.5\)},
        ymin=0,
        ymax=1,
        legend pos=south west,
    ]
        \addplot[color=DLRDarkerYellow,mark=triangle*] table[x=n,y=fixed_precision]{graphics/data/benchmark/halo_wide.dat};
        \addlegendentry{Fixed}
        \addplot[color=DLRDarkerBlue,mark=*] table[x=n,y=calibrated_sN_precision]{graphics/data/benchmark/halo_wide.dat};
        \addlegendentry{Calibrated}
    \end{axis}
    \autosafepairedplotbounds{40pt}{2pt}{34pt}{7pt}
\end{tikzpicture}%
        }%
        \caption{Precision of the represented region at \(\zeta = 0.5\) for the same anchor counts and parameterizations.}\label{fig:HaloAnchorCount-Precision}
    \end{subfigure}
    \caption{Measured exterior halo and precision as the number of anchor points increases; both horizontal axes use a logarithmic scale. The fixed-parameter halo grows from \num{1.027} to \num{3.426}, whereas the calibrated halo shrinks to \num{0.102}.}\label{fig:HaloAnchorCount}
\end{figure}

\FloatBarrier{}

\subsection{One-Class Baselines}
\Cref{tab:BaselineAUPR} compares the represented region with standard one-class estimators on synthetic domains with known membership.
The baselines are KDE, GMM selected using BIC, \(k\)-NN distance, OC-SVM, Isolation Forest, and the convex hull.
Given support vector data description yields, for kernels with constant self-similarity such as the RBF kernel, the same decision function as the OC-SVM with correspondingly chosen regularization, it is not separately evaluated~\cite{Schoelkopf2001, Tax2004}.
The comparison does not claim superior density estimation; it shows that the proposed assurance-oriented construction remains competitive on the two-dimensional suite while retaining its behavior on non-convex supports.
All two-dimensional domains use \(X = [-5, 5]^2\): the linear ODD satisfies \(x_2 \geq x_1 - 3\), the annulus ODD satisfies \(1 \leq x_1^2 + x_2^2 \leq 16\), the two-blob ODD is the union of disks of radius \(1.5\) centered at \((-2.5, -2.5)\) and \((2.5, 2.5)\), and the banana ODD satisfies \(\lvert x_2 - 0.4 x_1^2 + 2 \rvert \leq 1\).
The five-dimensional constrained ODD is the intersection of \(X = [-5, 5]^5\), \(x_2 \geq x_1 - 3\), and \(\sum_{k = 1}^{5} x_k^2 \leq 60\).
For each domain, \(\num{100000}\) validation points are sampled uniformly from the taxonomy box enlarged by a factor of two about its center.
KDE selects the greatest training log-likelihood among Scott's rule, Silverman's rule, and bandwidths \(0.05\), \(0.1\), \(0.2\), and \(0.5\).
The GMM selects \(1\), \(2\), \(4\), or \(8\) components by BIC\@.
The OC-SVM selects the greatest mean training decision margin over \(\nu \in \{0.01, 0.05, 0.1\}\) and kernel coefficients \(\gamma \in \{\text{scale}, 1, 10\}\).
The \(k\)-NN baseline uses \(k = 5\), Isolation Forest uses \(200\) estimators, and every experiment uses random seed \(42\).

\begin{table*}[htb]
    \caption{AUPR against the known region for \(N = \num{1000}\) anchors. autoSAFE ranks the validation points by the log-survival score \(-S(\bm{x})\). Values are rounded to three decimal places; boldface indicates the method with the highest AUPR in each dataset.}\label{tab:BaselineAUPR}
    \centering
    \begin{tblr}{
            colspec={l *{5}{Q[si={table-format=1.3},r]} c},
            colsep=4pt,
            row{1}={guard},
        }
        \toprule
        Method            & Linear          & Annulus         & Two Blobs       & Banana          & 5D ODD          & Order Deviation   \\
        \midrule
        autoSAFE          & 0.992           & 0.990           & 0.985           & 0.970           & 0.856           & \(\num{5.8e-15}\) \\
        KDE               & 0.996           & 0.995           & 0.996           & \bfseries 0.986 & 0.692           & ---               \\
        GMM (BIC)         & \bfseries 0.998 & \bfseries 0.998 & \bfseries 0.997 & 0.983           & \bfseries 0.882 & ---               \\
        \(k\)-NN distance & 0.991           & 0.991           & 0.984           & 0.970           & 0.854           & ---               \\
        OC-SVM            & 0.995           & 0.817           & 0.997           & 0.337           & 0.868           & ---               \\
        Isolation Forest  & 0.952           & 0.918           & 0.972           & 0.821           & 0.675           & ---               \\
        Convex hull       & 0.988           & 0.901           & 0.482           & 0.332           & 0.632           & ---               \\
        \bottomrule
    \end{tblr}
\end{table*}

\begin{table}[htb]
    \caption{Fixed-threshold comparison against the known non-convex regions for \(N = \num{1000}\) anchors. autoSAFE is evaluated at \(\zeta = 0.5\), while convex-hull membership is binary. The mean false-positive boundary distance is approximated in the original coordinate space by the distance from each false-positive validation point to its nearest validation point with the opposite true label; lower values are better.}\label{tab:BaselineFixedThreshold}
    \centering
    \begin{tblr}{
            colspec={l l *{3}{Q[si={table-format=1.3},r]}},
            colsep=5pt,
            row{1}={guard},
        }
        \toprule
        ODD       & Method      & IoU   & False-positive rate & Mean false-positive boundary distance \\
        \midrule
        Annulus   & autoSAFE    & 0.892 & 0.016               & 0.144                                 \\
        Annulus   & Convex hull & 0.898 & 0.009               & 0.358                                 \\
        Two blobs & autoSAFE    & 0.855 & 0.006               & 0.107                                 \\
        Two blobs & Convex hull & 0.488 & 0.037               & 1.043                                 \\
        Banana    & autoSAFE    & 0.775 & 0.012               & 0.113                                 \\
        Banana    & Convex hull & 0.338 & 0.080               & 1.311                                 \\
        \bottomrule
    \end{tblr}
\end{table}

The five-dimensional constrained ODD is the most challenging case of the suite: autoSAFE ranks third, behind GMM and OC-SVM, and its region at the fixed threshold of \(\zeta = 0.5\) strongly over-approximates the true region, with a false-positive rate of \num{0.864}, as the informative thresholds lie very close to one (cf.\ \cref{subsec:Limitations}).
The order-deviation value is the maximum absolute affinity difference across five permutations of the ID anchors and was measured only for autoSAFE; the dashes denote unavailable measurements for the other methods. 

\clearpage{}

\subsection{Anchor Count and Parameter Sensitivity}\label{subsec:ParameterSensitivity}
\begin{figure}[htb]
    \begin{subfigure}[t]{0.475\linewidth}
        \centering
        \resizebox{\linewidth}{!}{%
            \begin{tikzpicture}
    \pgfdeclarelayer{background}
    \pgfdeclarelayer{foreground}
    \pgfsetlayers{background,main,foreground}

    \begin{axis}[
        autosafe plot,
        xmode=log,
        log basis x=10,
        xmin=3,
        xmax=10000,
        xlabel={Anchor count \(N\)},
        ylabel={AUPR},
        ymin=0.5,
        ymax=1,
    ]
        \begin{pgfonlayer}{axis background}
            \addplot[name path=U1,draw=none,forget plot]
                table[x=n,y expr={\thisrow{aupr}+\thisrow{aupr_std}}]{graphics/data/benchmark/per_n.dat};
            \addplot[name path=L1,draw=none,forget plot]
                table[x=n,y expr={\thisrow{aupr}-\thisrow{aupr_std}}]{graphics/data/benchmark/per_n.dat};
            \addplot[color=DLRDarkBlue!80,opacity=0.2,forget plot]
                fill between[of=U1 and L1];
        \end{pgfonlayer}

        \begin{pgfonlayer}{axis foreground}
            \addplot[color=DLRDarkerBlue,mark=none]
                table[x=n,y=aupr]{graphics/data/benchmark/per_n.dat};
        \end{pgfonlayer}
    \end{axis}
    \autosafepairedplotbounds{48pt}{11pt}{34pt}{8pt}
\end{tikzpicture}%
        }%
        \caption{Mean AUPR against the known ODD as the number of anchors increases. The shaded region shows one standard deviation across five random seeds.}\label{fig:PerNValidation-AUPR}
    \end{subfigure}
    \hfill
    \begin{subfigure}[t]{0.475\linewidth}
        \centering
        \resizebox{\linewidth}{!}{%
            \begin{tikzpicture}
    \begin{axis}[
        autosafe plot,
        xmode=log,
        log basis x=10,
        ymode=log,
        xmin=3,
        xmax=10000,
        ymin=1e-4,
        ymax=1e3,
        xlabel={Anchor count \(N\)},
        ylabel={Precision-curve error \(1 - R^2\)},
    ]
        \addplot[color=DLRDarkerYellow,mark=none]
            table[x=n,y expr={1-\thisrow{r2_precision}}]{graphics/data/benchmark/per_n.dat};
    \end{axis}
    \autosafepairedplotbounds{48pt}{11pt}{34pt}{8pt}
\end{tikzpicture}%
        }%
        \caption{Disagreement, measured as \(1 - R^2\), between the precision--affinity-threshold curves obtained using the known ODD and the convex hull as references.}\label{fig:PerNValidation-Precision}
    \end{subfigure}
    \caption{AUPR and precision-curve agreement for the two-dimensional linear ODD as the number of anchors increases. The AUPR reaches \(\num{0.99902(11)}\) and the precision-curve agreement reaches \(R^2 = \num{0.99965}\) at \(N = \num{10000}\).}\label{fig:PerNValidation}
\end{figure}

\begin{figure}[htb]
    \begin{subfigure}[t]{0.475\linewidth}
        \centering
        \resizebox{\linewidth}{!}{%
            \begin{tikzpicture}
    \begin{axis}[
        autosafe sensitivity heatmap,
        point meta min=0.97,
        point meta max=1,
        colorbar,
        colorbar style={
            ylabel={AUPR},
            ytick={0.97,0.98,0.99,1},
            yticklabels={0.97,0.98,0.99,1.00},
        },
    ]
        \addplot[
            autosafe sensitivity cells,
        ]
            table[x expr={mod(\coordindex,4)},y expr={floor(\coordindex/4)},meta=aupr]{graphics/data/benchmark/sensitivity_linear_n1000.dat};
    \end{axis}
\end{tikzpicture}%
        }%
        \caption{The two-dimensional linear ODD is the intersection of \(X = [-5, 5]^2\) with the half-space \(x_2 \geq x_1 - 3\).}\label{fig:ParameterSensitivity-Linear}
    \end{subfigure}
    \hfill
    \begin{subfigure}[t]{0.475\linewidth}
        \centering
        \resizebox{\linewidth}{!}{%
            \begin{tikzpicture}
    \begin{axis}[
        autosafe sensitivity heatmap,
        point meta min=0.97,
        point meta max=1,
        colorbar,
        colorbar style={
            ylabel={AUPR},
            ytick={0.97,0.98,0.99,1},
            yticklabels={0.97,0.98,0.99,1.00},
        },
    ]
        \addplot[
            autosafe sensitivity cells,
        ]
            table[x expr={mod(\coordindex,4)},y expr={floor(\coordindex/4)},meta=aupr]{graphics/data/benchmark/sensitivity_annulus_n1000.dat};
    \end{axis}
\end{tikzpicture}%
        }%
        \caption{The two-dimensional annulus ODD is the ring \(1 \leq x_1^2 + x_2^2 \leq 16\) within \(X = [-5, 5]^2\), which excludes the central hole.}\label{fig:ParameterSensitivity-Annulus}
    \end{subfigure}
    \par\medskip
    \hfill
    \begin{subfigure}[t]{0.475\linewidth}
        \centering
        \resizebox{\linewidth}{!}{%
            \begin{tikzpicture}
    \begin{axis}[
        autosafe sensitivity heatmap,
        point meta min=0.8,
        point meta max=1,
        colorbar,
        colorbar style={
            ylabel={AUPR},
            ytick={0.8,0.85,0.9,0.95,1},
            yticklabels={0.80,0.85,0.90,0.95,1.00},
        },
    ]
        \addplot[
            autosafe sensitivity cells,
        ]
            table[x expr={mod(\coordindex,4)},y expr={floor(\coordindex/4)},meta=aupr]{graphics/data/benchmark/sensitivity_poly5d_n1000.dat};
    \end{axis}
\end{tikzpicture}%
        }%
        \caption{The five-dimensional constrained ODD is the intersection of \(X = [-5, 5]^5\), the half-space \(x_2 \geq x_1 - 3\), and the ball \(\sum_{k = 1}^{5} x_k^2 \leq 60\).}\label{fig:ParameterSensitivity-Polytope}
    \end{subfigure}
    \hfill\null
    \caption{AUPR sensitivity to the dimensionless calibration constants \(s\) and \(\gamma{}\) at \(N = \num{1000}\). Across the full grid, the AUPR remains within \numrange{0.978}{0.998} for the linear ODD and within \numrange{0.973}{0.998} for the annulus ODD\@. For the same anchor count, the five-dimensional constrained ODD ranges over \numrange{0.825}{0.909} and reaches its maximum at \(\gamma = \num{0.5}\) and \(s = 1\). The scheduling rule \(s_N \approx \num{2.63}\) at this anchor count yields AUPR values of \numrange{0.986}{0.994}, \numrange{0.984}{0.995}, and \numrange{0.846}{0.860} for the three ODDs.}\label{fig:ParameterSensitivity}
\end{figure}

The sensitivity of the sparse five-dimensional ODD to \(s\) suggests that the width multiple should decrease with the dimension \(n\): the number of anchors within a fixed multiple of \(\tilde{d}\) grows rapidly with \(n\), and with it the accumulated noisy-OR tail at any fixed \(s\).
A dimension-dependent constant \(c_n > 0\) in the scheduling rule, \(s_{n,N} = c_n s_N\), leaves \cref{thm:AsympODDSimCalibrated} unchanged because \(n\) is fixed as \(N\) grows.
For \(c_n = n^{-1/2}\) and \(\gamma = 1\), the IoU of the five-dimensional ODD at \(\zeta = 0.5\) rises from \numrange{0.022}{0.029} without the correction factor \(c_n\) to \numrange{0.217}{0.289} with the correction factor \(c_n\) for anchor counts from \num{100} to \num{10000}, and its AUPR at \(N = \num{1000}\) from \num{0.850} to \num{0.906}.
On the two-dimensional ODDs at \(N = \num{1000}\), the IoU changes from \num{0.894} and \num{0.870} to \num{0.901} and \num{0.917}, whereas at \(N = 100\) the annulus IoU decreases from \num{0.772} to \num{0.679}.
A stronger scaling of \(c_n = n^{-1}\) narrows the two-dimensional kernels excessively, with IoU values of \num{0.709} and \num{0.772} at \(N = \num{1000}\).
Since even \(c_n = n^{-1/2}\) remains below the five-dimensional IoU of \num{0.645} reached at \(s = 0.75\), a principled and justified choice of \(c_n\) is left to future work.

\FloatBarrier{}

\subsection{Adaptive Threshold Grid}\label{subsec:ThresholdGrid}
Let \(L \geq 2\) be the requested number of threshold pairs and let \(S = \ln(1 - \zeta)\) be the corresponding log-survival threshold.
The grid always contains the endpoint pairs \((\zeta, S) = (0, 0)\) and \((1, -\infty)\).
Approximately one quarter of the remaining \(L - 2\) entries are assigned to a lower logarithmic edge, one quarter to a linear middle interval, and one half to an upper logarithmic edge.
For the lower edge, \(t\) is spaced linearly from \(1\) to \(d_{\mathrm{lo}}\) and \(\zeta = 10^{-t}\), where \(d_{\mathrm{lo}}\) is the larger of \(16\) and \(-\log_{10} \zeta_{\min}^+\) for the smallest positive observed affinity \(\zeta_{\min}^+\).
The middle section is linear in \(\zeta{}\) from \num{0.1} to \num{0.9}.
For the upper edge, \(t\) is spaced linearly from \(1\) to \(d_{\mathrm{hi}}\), \(S = -t \ln 10\), and \(\zeta = 1 - \exp(S)\), where \(d_{\mathrm{hi}}\) is the larger of \(16\) and \(\lvert S_{\min}\rvert/\ln 10\) for the most negative finite observed log-survival.
Duplicate junction pairs are removed, and all decisions use the stored value of \(S\) directly through \cref{eq:LogThreshold}.

\FloatBarrier{}

\subsection{Distribution-Free Calibration of the Membership Threshold}\label{subsec:ConformalThreshold}
Let \(Q(\bm{x}) := -S(\bm{x}) = -\ln(1 - \alpha(\bm{x}))\) be the log-space affinity score and let \(Q_{(1)} \leq \cdots \leq Q_{(N_{\mathrm{cal}})}\) be the ordered scores of a held-out ID calibration set of size \(N_{\mathrm{cal}}\) that is exchangeable with a future ID sample.
For a target false-exclusion budget \(\varepsilon \in (0, 1)\), define \(k := \lfloor \varepsilon(N_{\mathrm{cal}} + 1)\rfloor{}\), \(\hat{t} := Q_{(k)}\), and \(\hat{\zeta} := 1 - \exp(-\hat{t})\).
The construction is certifiable only when \(k \geq 1\), which requires \(N_{\mathrm{cal}} \geq 1/\varepsilon - 1\).
By the split-conformal rank argument~\cite{Angelopoulos2023},
\begin{align}
    \mathbb{P}\left(\alpha(\bm{x}_{N_{\mathrm{cal}} + 1}) < \hat{\zeta}\right) = \mathbb{P}\left(Q(\bm{x}_{N_{\mathrm{cal}} + 1}) < \hat{t}\right) \leq \frac{k}{N_{\mathrm{cal}} + 1} \leq \varepsilon\text{.}
\end{align}
The calibration set is reserved before anchor de-duplication, is disjoint from the anchor set, and retains the operational record frequencies required for exchangeability with a future ID record.
Its scores are computed using the final representation after anchor de-duplication and any OOD adjustment.
The required ordering \(\xi < \hat{\zeta}\) is checked explicitly when the two thresholds are used together.
The conformal guarantee controls false exclusion only; false inclusion is addressed separately by the outer-halo analysis and the observed-OOD consistency constraint in \cref{sec:AsympSimProof,def:OODConsCons}.
Calibrating on \(Q\) preserves the ordering of \(\alpha{}\) while retaining distinctions that finite-precision affinity values can lose near one.
For each of the linear, annular, and five-dimensional constrained domains, total ID pools of size \(N_{\mathrm{pool}} \in \{500, 1000, 5000\}\) were split equally into \(N = N_{\mathrm{cal}} \in \{250, 500, 2500\}\) anchors and calibration records, with fresh sets of \num{5000} ID and \num{5000} out-of-region test samples per repetition.
Across 20 repetitions, the observed false-exclusion-rate ranges are \numrange{0.0063}{0.0110}, \numrange{0.0493}{0.0523}, and \numrange{0.0985}{0.1035} for \(\varepsilon = \num{0.01}\), \num{0.05}, and \num{0.10}, respectively, while \qtyrange{93.6}{99.8}{\percent} of out-of-region points remain excluded.

\FloatBarrier{}

\subsection{Non-Convex Exclusion and Anchor-Set Robustness}
The excluded VCAS region is defined by \(\lvert h \rvert \leq \qty{300}{\foot}\) and \(\tau \leq \qty{8}{\second}\); it represents low vertical separation shortly before the closest point of approach (CPA).
\begin{table}[htb]
    \caption{False-positive rate on held-out points from the excluded VCAS region. Relative changes use autoSAFE without OOD constraints as the baseline; lower values are better.}\label{tab:HeldOutHole}
    \centering
    \begin{tblr}{
            colspec={l Q[si={table-format=1.4},r] l},
            row{1}={guard},
        }
        \toprule
        Method                        & False-positive rate & Relative change from autoSAFE    \\
        \midrule
        autoSAFE                      & 0.0420              & \mbox{\textemdash}               \\
        autoSAFE with OOD constraints & 0.0055              & \qty{86.9}{\percent} improvement \\
        Convex hull                   & 0.5470              & \qty{1202.4}{\percent} worsening \\
        \bottomrule
    \end{tblr}
\end{table}

The sampling-density and resolution-cell de-duplication probes are shown in \cref{fig:DuplicateSensitivity}.
Duplicates affect the calibrated construction in two ways: an anchor present \(m > 1\) times raises its local contribution from \(\alpha_i\) to \(1 - (1 - \alpha_i)^m > \alpha_i\) away from the anchor itself, which the probe reproduces exactly, and repeated records drive nearest-neighbor distances toward zero, which shifts the median distance \(\tilde{d}\) and thereby every calibrated kernel width.
On the annulus, one additional copy of \qty{10}{\percent} of the anchors already lowers the IoU with the duplicate-free region to \num{0.971}; in the de-duplication probe, duplicates lower it to \num{0.685} for the calibrated construction, compared with \num{0.820} for fixed parameters, while resolution-cell de-duplication restores it to \num{0.936}.

\begin{figure}[htb]
    \begin{subfigure}[t]{0.475\linewidth}
        \centering
        \resizebox{\linewidth}{!}{%
            \begin{tikzpicture}
    \begin{axis}[
        autosafe plot,
        xmode=log,
        log basis x=10,
        xtickten={2,3},
        minor xtick={200,300,400,500,600,700,800,900,2000,3000},
        xmin=100,
        xmax=3000,
        xlabel={Anchor count \(N\)},
        ylabel={IoU with true region},
        ymin=0.6,
        ymax=1,
        legend pos=south east,
    ]
        \addplot[color=DLRDarkerBlue,mark=*] table[x=n,y=annulus]{graphics/data/benchmark/density.dat};
        \addlegendentry{Annulus}
        \addplot[color=DLRDarkerGreen,mark=square*] table[x=n,y=linear]{graphics/data/benchmark/density.dat};
        \addlegendentry{Linear}
    \end{axis}
    \autosafepairedplotbounds{40pt}{13pt}{34pt}{7pt}
\end{tikzpicture}%
        }%
        \caption{IoU with the known annulus and linear ODDs as the number of anchors increases.}\label{fig:DuplicateSensitivity-Density}
    \end{subfigure}
    \hfill
    \begin{subfigure}[t]{0.475\linewidth}
        \centering
        \resizebox{\linewidth}{!}{%
            \begin{tikzpicture}
    \begin{axis}[
        autosafe plot,
        xmin=0,
        xmax=0.05,
        xlabel={Resolution-cell width},
        ylabel={IoU with duplicate-free region},
        ymin=0.6,
        ymax=1,
        xtick={0,0.01,0.05},
        xticklabels={0,0.01,0.05},
        legend pos=south east,
    ]
        \addplot[color=DLRDarkerBlue,mark=*] table[x=tol,y=annulus]{graphics/data/benchmark/dedup.dat};
        \addlegendentry{Annulus}
        \addplot[color=DLRDarkerGreen,mark=square*] table[x=tol,y=linear]{graphics/data/benchmark/dedup.dat};
        \addlegendentry{Linear}
    \end{axis}
    \autosafepairedplotbounds{40pt}{13pt}{34pt}{7pt}
\end{tikzpicture}%
        }%
        \caption{IoU with the duplicate-free reference regions as the candidate resolution-cell width increases.}\label{fig:DuplicateSensitivity-Deduplication}
    \end{subfigure}
    \caption{Effect of sampling density and resolution-cell de-duplication. Increasing the density of distinct anchors raises the IoU with the known region from approximately \numrange{0.75}{0.78} to \num{0.91}; in the duplicate probe, collapsing records within shared resolution cells restores the IoU with the duplicate-free region from \num{0.685} to \num{0.936}. These results motivate the deterministic ingestion rule in \cref{alg:autoSAFE}: record multiplicity is retained as coverage and provenance metadata but does not instantiate additional kernels.}\label{fig:DuplicateSensitivity}
\end{figure}

\begin{figure}[htb]
    \centering
    \begin{minipage}{0.475\linewidth}
        \centering
        \resizebox{\linewidth}{!}{%
            \begin{tikzpicture}
    \pgfdeclarelayer{background}
    \pgfdeclarelayer{foreground}
    \pgfsetlayers{background,main,foreground}

    \begin{axis}[
        autosafe plot,
        xmode=log,
        log basis x=10,
        xtickten={2,3},
        minor xtick={200,300,400,500,600,700,800,900,2000,3000},
        xmin=100,
        xmax=3000,
        ymin=0.75,
        ymax=1,
        xlabel={Subset size \(N\)},
        ylabel={Pairwise region IoU},
        legend pos=south east,
    ]
        \begin{pgfonlayer}{axis background}
            \addplot[name path=LinearUpper,draw=none,forget plot]
                table[x=n,y=linear_max]{graphics/data/benchmark/subset.dat};
            \addplot[name path=LinearLower,draw=none,forget plot]
                table[x=n,y=linear_min]{graphics/data/benchmark/subset.dat};
            \addplot[color=DLRDarkerBlue,opacity=0.2,forget plot]
                fill between[of=LinearUpper and LinearLower];
            \addplot[name path=AnnulusUpper,draw=none,forget plot]
                table[x=n,y=annulus_max]{graphics/data/benchmark/subset.dat};
            \addplot[name path=AnnulusLower,draw=none,forget plot]
                table[x=n,y=annulus_min]{graphics/data/benchmark/subset.dat};
            \addplot[color=DLRDarkerGreen,opacity=0.2,forget plot]
                fill between[of=AnnulusUpper and AnnulusLower];
        \end{pgfonlayer}

        \begin{pgfonlayer}{axis foreground}
            \addplot[color=DLRDarkerBlue,mark=none]
                table[x=n,y=linear_mean]{graphics/data/benchmark/subset.dat};
            \addlegendentry{Linear}
            \addplot[color=DLRDarkerGreen,mark=none]
                table[x=n,y=annulus_mean]{graphics/data/benchmark/subset.dat};
            \addlegendentry{Annulus}
        \end{pgfonlayer}
    \end{axis}
\end{tikzpicture}%
        }%
    \end{minipage}
    \caption{Mean pairwise region IoU across ten independently selected anchor subsets as the subset size increases; the shaded regions span the minimum and maximum of the \num{45} pairwise comparisons at each size. The mean rises from \numrange{0.77}{0.79} at \(N = 100\) to approximately \num{0.96} at \(N = 3000\), where every pair exceeds \num{0.95}. The narrowing range shows that the represented region becomes less sensitive to which anchors are selected as the subset grows.}\label{fig:SubsetStability}
\end{figure}

The calibrated full-data aviation evaluations used all \num{622110} VCAS anchors and all \num{385925} HCAS anchors, with \num{e6} evaluation samples for each use case.
At the smallest positive evaluated affinity threshold, precision and recall against the known ODD were \num{1.000} and \num{0.633} for VCAS and \num{0.995} and \num{0.655} for HCAS, compared with prevalences of \num{0.789} and \num{0.738} in the respective evaluation samples.
Because the recorded operational states do not fill the known ODDs, recall below one reflects regions of the known ODD that no anchor point represents rather than a reconstruction error of \(\mathcal{O}\).
\begin{figure}[htb]
    \hspace*{\fill}
    \begin{subfigure}[t]{0.475\linewidth}
        \centering
        \resizebox{\linewidth}{!}{%
            \begin{tikzpicture}
    \begin{axis}[
        autosafe plot,
        name=main,
        xmode=log,
        x dir=reverse,
        log basis x=10,
        xlabel={\(1 - \zeta{}\)},
        ylabel={Score},
        xtick={1,1e-1,1e-2,1e-3,1e-4},
        ytick={0,0.5,1.0},
        xmin=0.0001,
        xmax=1,
        ymin=0,
        ymax=1.05,
        legend pos=south west,
    ]
        \addplot [mark=none, color=DLRDarkerBlue, line width=1.0pt]
        table [col sep=comma, x=one_minus_zeta, y=precision] {graphics/data/vcas_box_sweep.csv};
        \addlegendentry{Precision}

        \addplot [mark=none, color=DLRDarkerGreen, line width=1.0pt]
        table [col sep=comma, x=one_minus_zeta, y=recall] {graphics/data/vcas_box_sweep.csv};
        \addlegendentry{Recall}

        \addplot [mark=none, color=DLRGrey, dashed, line width=1.0pt]
        table [col sep=comma, x=one_minus_zeta, y=prevalence] {graphics/data/vcas_box_sweep.csv};
        \addlegendentry{Prevalence}

        \draw [DLRGrey, dotted, line width=0.8pt] (axis cs:0.5,0) -- (axis cs:0.5,1.05)
            node [pos=0.87, anchor=west, font=\scriptsize, inner sep=1pt] {\(\zeta = 0.5\)};
        \draw [DLRGrey, dotted, line width=0.8pt] (axis cs:0.00767304,0) -- (axis cs:0.00767304,1.05)
            node [pos=0.87, anchor=west, font=\scriptsize, inner sep=1pt] {\(\hat{\zeta}_{0.1}\)};
    \end{axis}
    \begin{axis}[
        autosafe plot,
        at={(main.south west)},
        anchor=north west,
        yshift=-3.8em,
        height=1pt,
        axis y line=none,
        axis x line*=bottom,
        xmode=log,
        x dir=reverse,
        xmin=0.0001,
        xmax=1,
        ymin=0,
        ymax=1,
        xtick={1,1e-1,1e-2,1e-3,1e-4},
        xticklabels={\(0\),\(0.9\),\(0.99\),\(0.999\),\(0.9999\)},
        xlabel={\(\zeta{}\)},
    ]
    \end{axis}
\end{tikzpicture}%
        }%
        \caption{Precision and recall of the VCAS ODD derived from all \num{622110} anchor points relative to the affinity threshold. Precision stays above \num{0.999} at every threshold, while recall decreases from \num{0.633} to \num{0.558} at \(1 - \zeta = \num{e-4}\).}\label{fig:CAS-PR-VCAS}
    \end{subfigure}
    \hfill
    \begin{subfigure}[t]{0.475\linewidth}
        \centering
        \resizebox{\linewidth}{!}{%
            \begin{tikzpicture}
    \begin{axis}[
        autosafe plot,
        name=main,
        xmode=log,
        x dir=reverse,
        log basis x=10,
        xlabel={\(1 - \zeta{}\)},
        ylabel={Score},
        xtick={1,1e-1,1e-2,1e-3,1e-4},
        ytick={0,0.5,1.0},
        xmin=0.0001,
        xmax=1,
        ymin=0,
        ymax=1.05,
        legend pos=south west,
    ]
        \addplot [mark=none, color=DLRDarkerBlue, line width=1.0pt]
        table [col sep=comma, x=one_minus_zeta, y=precision] {graphics/data/hcas_box_sweep.csv};
        \addlegendentry{Precision}

        \addplot [mark=none, color=DLRDarkerGreen, line width=1.0pt]
        table [col sep=comma, x=one_minus_zeta, y=recall] {graphics/data/hcas_box_sweep.csv};
        \addlegendentry{Recall}

        \addplot [mark=none, color=DLRGrey, dashed, line width=1.0pt]
        table [col sep=comma, x=one_minus_zeta, y=prevalence] {graphics/data/hcas_box_sweep.csv};
        \addlegendentry{Prevalence}

        \draw [DLRGrey, dotted, line width=0.8pt] (axis cs:0.5,0) -- (axis cs:0.5,1.05)
            node [pos=0.87, anchor=west, font=\scriptsize, inner sep=1pt] {\(\zeta = 0.5\)};
        \draw [DLRGrey, dotted, line width=0.8pt] (axis cs:0.0139075,0) -- (axis cs:0.0139075,1.05)
            node [pos=0.87, anchor=west, font=\scriptsize, inner sep=1pt] {\(\hat{\zeta}_{0.1}\)};
    \end{axis}
    \begin{axis}[
        autosafe plot,
        at={(main.south west)},
        anchor=north west,
        yshift=-3.8em,
        height=1pt,
        axis y line=none,
        axis x line*=bottom,
        xmode=log,
        x dir=reverse,
        xmin=0.0001,
        xmax=1,
        ymin=0,
        ymax=1,
        xtick={1,1e-1,1e-2,1e-3,1e-4},
        xticklabels={\(0\),\(0.9\),\(0.99\),\(0.999\),\(0.9999\)},
        xlabel={\(\zeta{}\)},
    ]
    \end{axis}
\end{tikzpicture}%
        }%
        \caption{Precision and recall of the HCAS ODD derived from all \num{385925} anchor points relative to the affinity threshold. Precision stays above \num{0.99} at every threshold, while recall decreases from \num{0.655} to \num{0.406} at \(1 - \zeta = \num{e-4}\).}\label{fig:CAS-PR-HCAS}
    \end{subfigure}
    \hspace*{\fill}
    \caption{Precision and recall of the data-driven ODD \(\mathcal{O}\) against the known ODDs of VCAS and HCAS~\cite{Julian2019} as the affinity threshold \(\zeta{}\) approaches one, with \(1 - \zeta{}\) on a logarithmic scale above the corresponding \(\zeta{}\). The dotted lines mark \(\zeta = 0.5\) and the split-conformal threshold \(\hat{\zeta}_{0.1}\), and the dashed line marks the share of evaluation samples inside the known ODD. Recall below one reflects regions of the known ODD that the recorded states never visit.}\label{fig:CAS-PR}
\end{figure}
The full-data aviation runs additionally compare the autoSAFE membership decisions with seven reference regions fitted to the same anchors, including the core support from density-based spatial clustering of applications with noise (DBSCAN)~\cite{Ester1996}.
Consequently, the IoU values in \cref{tab:AviationReferenceAgreement} quantify agreement between representations and are not baseline-versus-ground-truth accuracy measurements.
For the production-scale data, the global envelope identified as \texttt{hull\_single} by the evaluation artifact is a radial approximation based on the mean and maximum anchor radius, rather than an exact convex hull.

\begin{table}[htb]
    \caption{IoU between autoSAFE and anchor-derived reference regions at the smallest positive evaluated affinity threshold, \(\zeta = \num{1e-16}\), using \num{e6} evaluation samples per use case.}\label{tab:AviationReferenceAgreement}
    \centering
    \begin{tblr}{
            colspec={l *{2}{Q[si={table-format=1.3},r]}},
            row{1}={guard},
        }
        \toprule
        Reference region              & VCAS  & HCAS  \\
        \midrule
        Global radial envelope        & 0.522 & 0.550 \\
        Clustered hull reference      & 0.534 & 0.657 \\
        \(k\)-NN support              & 0.737 & 0.539 \\
        \(k\)-means boundaries        & 0.684 & 0.762 \\
        KDE superlevel set            & 0.250 & 0.174 \\
        Clustered KDE superlevel sets & 0.250 & 0.184 \\
        DBSCAN core support           & 0.786 & 0.573 \\
        \bottomrule
    \end{tblr}
\end{table}

For the VCAS subset experiment, two independently seeded sets of \num{5000} anchors were selected from the complete anchor pool and evaluated on independently sampled sets of \num{200000} points.
The known VCAS ODD is the only reference in this experiment that does not itself change with the selected anchor subset.
As shown in \cref{tab:AviationSubsetStability}, the precision and IoU of both subsets remain within \num{0.002} of the full-data evaluation despite the approximately \(124\)-fold reduction in anchor count.

\begin{table*}[htb]
    \caption{Agreement with the known VCAS ODD at the smallest positive evaluated affinity threshold. The subset runs use independently drawn anchor and evaluation sets; the values therefore measure end-to-end stability rather than paired pointwise agreement.}\label{tab:AviationSubsetStability}
    \centering
    \small
    \begin{tblr}{
            colspec={l *{2}{Q[si={table-format=7.0},r]} *{4}{Q[si={table-format=1.3},r]}},
            colsep=2pt,
            row{1}={guard},
        }
        \toprule
        Run                    & Anchors & Evaluation samples & Prevalence     & Precision & Recall & IoU   \\
        \midrule
        VCAS, subset seed 1    & 5000    & 200000             & 0.789          & 0.998     & 0.633  & 0.632 \\
        VCAS, subset seed 2    & 5000    & 200000             & 0.789          & 0.998     & 0.634  & 0.633 \\
        VCAS, complete dataset & 622110  & 1000000            & 0.789          & 1.000     & 0.633  & 0.633 \\
        \bottomrule
    \end{tblr}
\end{table*}

\FloatBarrier{}

\subsection{Permutation Stability}
Across 20 permutations of the ID and OOD sample orderings, the derived covariance matrices were identical, while the evaluated affinities differed by at most \num{1.3e-15}; the residual stems solely from the finite-precision summation order of the batched evaluation (cf.\ \cref{subsec:NumericalAffinity}).

\FloatBarrier{}

\section{High-Dimensional ODD Representation}\label{sec:ODD_Representation}
To model the ODD structure \(\mathcal{O}\), especially in higher dimensions, a robust framework is required.
To model the parameter space \(X\), polytopes have proven effective~\cite{Nenchev2025}.
In general, given the complexity of the real world, the ODD polytope is not convex.
While polytopes can be concave, modeling the ODD polytope as a union of convex polytopes is helpful, enabling the use of convex optimization tools on each component polytope.
Thus, the ODD polytope \(\mathcal{P}\) can be defined as the union of a set of \(J\) unique convex non-intersecting polytopes \(\mathcal{C}_j\) as
\begin{align}
    \mathcal{P} = \bigcup_{j = 1}^J \mathcal{C}_j\text{,}
\end{align}
with the half-space definition of \(\mathcal{C}\) using the componentwise inequality
\begin{align}
    \mathcal{C}_j = \left\{\bm{x} \in \mathbb{R}^n : \bm{A}_j \bm{x} \leq \bm{b}_j\right\}\text{,}
\end{align}
where \(\bm{A}_j\) is the matrix containing the normal vectors to the \(j\)-th hyperplane defining the half-space and \(\bm{b}_j\) is the right-hand side vector that determines the distance of the hyperplane from the origin.

\section{Proof of Termination}\label{sec:ProofofTermination}
\begin{proposition}[Termination of Global OOD Consistency Adjustment]\label{prop:OODTermination}
    The global OOD consistency adjustment procedure in \cref{alg:autoSAFE} terminates in a finite number of steps, ensuring \(\alpha(\bm{x}) \leq \xi{}\) for all \(\bm{x} \in \mathcal{D}_\mathrm{OOD}\).
\end{proposition}

\begin{proof}
    If \(\mathcal{D}_\mathrm{OOD} = \emptyset{}\), the claim is immediate.
    Otherwise, let \(\bm{\Sigma}_i^{(0)}\) be the initial covariance of kernel \(i\), and define
    \begin{align}
        q_i^{\min} := \min_{\bm{z} \in \mathcal{D}_\mathrm{OOD}}
        (\bm{z} - \bm{x}_i)^\top
        \left(\bm{\Sigma}_i^{(0)}\right)^{-1}
        (\bm{z} - \bm{x}_i) > 0\text{.}
    \end{align}
    The minimum is positive because the OOD set is finite, the covariance is positive-definite, and \(\mathcal{A} \cap \mathcal{D}_\mathrm{OOD} = \emptyset{}\), which follows from \(\mathcal{A} \subseteq \mathcal{D}_\mathrm{ID}\) and \(\mathcal{D}_\mathrm{ID} \cap \mathcal{D}_\mathrm{OOD} = \emptyset{}\).

    Let \(t_i\) denote the number of times kernel \(i\) has been scaled.
    Then \(\bm{\Sigma}_i^{(t_i)} = c^{t_i} \bm{\Sigma}_i^{(0)}\), and for every \(\bm{z} \in \mathcal{D}_\mathrm{OOD}\),
    \begin{align}
        \alpha_i^{(t_i)}(\bm{z})
         & = \exp\left(-\frac{c^{-t_i}}{2}
        (\bm{z} - \bm{x}_i)^\top
        \left(\bm{\Sigma}_i^{(0)}\right)^{-1}
        (\bm{z} - \bm{x}_i)\right)                                      \\
         & \leq \exp\left(-\frac{q_i^{\min} c^{-t_i}}{2}\right)\text{.}
    \end{align}
    Set \(\xi_N := 1 - (1 - \xi)^{1/N}\), and define the finite per-kernel bound
    \begin{align}
        K_i := \max\left\{0,
        \left\lceil
        \frac{\ln\left(\frac{-2\ln(\xi_N)}{q_i^{\min}}\right)}{\ln\left(\frac{1}{c}\right)}
        \right\rceil\right\}\text{.}
        \label{eq:ood_iteration_bound}
    \end{align}
    After \(K_i\) selections of kernel \(i\), its local affinity is at most \(\xi_N\) at every OOD point.

    It remains to show that the algorithm cannot select kernel \(i\) more than \(K_i\) times.
    Whenever the loop has not terminated, its selected OOD point \(\bm{z}^*\) satisfies \(\alpha(\bm{z}^*) > \xi{}\).
    If all local affinities at \(\bm{z}^*\) were at most \(\xi_N\), then
    \begin{align}
        \alpha(\bm{z}^*)
        = 1 - \prod_{i = 1}^N \left(1 - \alpha_i(\bm{z}^*)\right)
        \leq 1 - (1 - \xi_N)^N = \xi\text{,}
    \end{align}
    a contradiction.
    Thus some local affinity at \(\bm{z}^*\) exceeds \(\xi_N\), and the dominant kernel selected by the algorithm also exceeds \(\xi_N\).
    A kernel that has already been selected \(K_i\) times cannot satisfy this condition and is never selected again while a violation remains.
    Consequently, each kernel is selected at most \(K_i\) times, and the total number of adjustments satisfies
    \begin{align}
        N_{\mathrm{adj}} \leq \sum_{i = 1}^N K_i < \infty\text{.}
    \end{align}
    This argument is unaffected by ties: any maximizer of either selection step satisfies the same bound.
    The loop therefore terminates after finitely many adjustments, and its stopping condition gives \(\alpha(\bm{z}) \leq \xi{}\) for every \(\bm{z} \in \mathcal{D}_\mathrm{OOD}\).
\end{proof}

\section{Asymptotic Similarity of the Kernel-Based ODD}\label{sec:AsympSimProof}
This section distinguishes the behavior of fixed kernel parameters from that of the calibrated parameters in \cref{eq:sigma_calibrated}.
For a set \(E \subseteq \mathbb{R}^n\), \(d(\bm{x}, E) := \inf_{\bm{y} \in E}\|\bm{x} - \bm{y}\|_2\) denotes the Euclidean point-to-set distance, \(B(\bm{x}, r)\) the open Euclidean ball, \(\partial E\) the boundary, and \(\interior(E)\) the interior.

\subsection{Proof of Scale Invariance}\label{sec:ScaleInvarianceProof}
This subsection proves \cref{prop:ScaleInvariance}.

\paragraph{Equivariance condition.}
Let \(T(\bm{x}) = \bm{A} \bm{x} + \bm{b}\) with \(\bm{A} = \diag(a_1, \dots, a_n)\), \(a_k > 0\), and let \(\alpha'\) denote the affinity built on the transformed anchor points \(\{T(\bm{x}_i)\}{}\) with covariance matrices \(\bm{\Sigma}_i'\).
The Mahalanobis form satisfies
\begin{align}
     & \left(T(\bm{x}) - T(\bm{x}_i)\right)^\top
    \left(\bm{A} \bm{\Sigma}_i \bm{A}^{\top}\right)^{-1}
    \left(T(\bm{x}) - T(\bm{x}_i)\right)
    \nonumber                                    \\
    =
     & \left(\bm{x} - \bm{x}_i\right)^\top
    \bm{\Sigma}_i^{-1}
    \left(\bm{x} - \bm{x}_i\right)\text{.}
\end{align}
Consequently, \(\alpha'(T(\bm{x})) = \alpha(\bm{x})\) for all \(\bm{x}\) if \(\bm{\Sigma}_i' = \bm{A} \bm{\Sigma}_i \bm{A}^{\top}\), i.e., if \(\sigma_{kk}'^{(i)} = a_k^2 \sigma_{kk}^{(i)}\) for all \(i\) and \(k\).

\paragraph{Similarity transformations.}
Let \(a_1 = \cdots = a_n = a > 0\).
Both nearest-neighbor modes are preserved under \(T\), and the distances satisfy \(d_i^{*\prime} = a d_i^*\), \(d_{ik}^{*\prime} = a d_{ik}^*\), and \(\tilde{d}' = a \tilde{d}\).
The calibrated parameters of \cref{eq:sigma_calibrated} consequently satisfy \(\eta' = \eta / a\), \(\kappa' = a^2 \kappa{}\), and \(\lambda' = a^2 \lambda{}\).
It follows that
\begin{align}
    \sigma_{kk}'^{(i)} = \left(a^2 \kappa - a^2 \lambda\right) \exp\left(-\frac{\eta}{a} a d_{ik}^*\right) + a^2 \lambda = a^2 \sigma_{kk}^{(i)}\text{,}
\end{align}
which is precisely the equivariance condition above.
The relative specification \(\lambda = \lambda_{\mathrm{rel}} \kappa{}\) is essential because an absolute lower bound would not transform as a squared length.

\paragraph{Per-dimension affine re-expressions.}
For general \(\bm{A} = \diag(a_1, \dots, a_n)\), the Euclidean nearest-neighbor assignment may change, so the preceding argument does not apply directly.
Assume that the complete derivation pipeline first normalizes every parameter by a per-dimension affine transform whose defining bounds re-express equivariantly with the data.
Re-expressing the raw data by \(T\) re-expresses the normalization bounds identically, so the normalized anchor set and every quantity derived from it remain unchanged.
The normalization therefore makes the complete pipeline invariant under arbitrary per-dimension affine re-expressions of the raw data.

\FloatBarrier{}

\subsection{Fixed-Parameter Behavior}
\begin{proposition}[Fixed-parameter behavior]\label{prop:AsympODDSim}
    Let the reference membership region \(R^* \subset \mathbb{R}^n\) be compact with positive finite volume and a boundary of Lebesgue measure zero.
    Let \(\bm{x}_1, \bm{x}_2, \ldots{}\) be independent and uniformly distributed on \(R^*\), let \(\alpha_N\) be the global affinity constructed from the first \(N\) anchors through \cref{eq:sigma_kk_mod}, and, for \(N \geq 2\), let \(R_N := \{\bm{x} : \alpha_N(\bm{x}) \geq \zeta{}\}{}\), where \(\zeta \in (0, 1)\), \(\kappa > 0\), \(\eta \geq 0\), and \(0 < \lambda \leq \kappa{}\) are held fixed as \(N\) grows.
    Then, almost surely, Lebesgue-almost every \(\bm{x} \in \interior(R^*)\) belongs to \(R_N\) for all sufficiently large \(N\).
    Moreover, for every \(N \geq 2\),
    \begin{align}
        R_N \subseteq (R^*)^{+\rho_N},
        \qquad
        \rho_N := \sqrt{2\kappa \ln\left(\frac{N}{\zeta}\right)}\text{,}
        \label{eq:fixed_bandwidth_halo}
    \end{align}
    where \((R^*)^{+r} := \{\bm{x} : d(\bm{x}, R^*) \leq r\}{}\) denotes the closed outward dilation of \(R^*\) by radius \(r\).
\end{proposition}

Here, \enquote{almost surely} means with probability one over the random sequence of anchor points; it does not express uncertainty about which conclusion holds once that sequence is fixed.

\begin{proof}
    Fix \(\bm{x} \in \interior(R^*)\) and choose \(\varepsilon > 0\) such that \(B(\bm{x}, \varepsilon) \subset R^*\) and \(\exp(-\varepsilon^2/(2\lambda)) \geq \zeta{}\).
    With probability one, an infinite i.i.d.\ sample from \(R^*\) hits this positive-volume ball; from the first such hit onward there is an anchor \(\bm{x}_i\) with \(\|\bm{x} - \bm{x}_i\|_2 \leq \varepsilon{}\).
    Since every diagonal entry of \(\bm{\Sigma}_i\) is at least \(\lambda{}\),
    \begin{align}
        \alpha_i(\bm{x})
        \geq \exp\left(-\frac{\|\bm{x} - \bm{x}_i\|_2^2}{2\lambda}\right)
        \geq \zeta\text{,}
    \end{align}
    and hence \(\alpha_N(\bm{x}) \geq \alpha_i(\bm{x}) \geq \zeta{}\) for every subsequent \(N\).
    Applying Fubini's theorem~\cite{Evans2025} to the sampling event and using the null boundary gives the first claim simultaneously for Lebesgue-almost every interior point.

    For the outer bound, fix \(\bm{x}\) at distance \(\delta = d(\bm{x}, R^*)\) from \(R^*\).
    Every anchor is at least \(\delta{}\) away, and every covariance eigenvalue is at most \(\kappa{}\), so
    \begin{align}
        \alpha_i(\bm{x})
        \leq \exp\left(-\frac{\delta^2}{2\kappa}\right)\text{.}
    \end{align}
    The union bound for the noisy-OR aggregation gives
    \begin{align}
        \alpha_N(\bm{x})
        \leq \sum_{i = 1}^{N} \alpha_i(\bm{x})
        \leq N \exp\left(-\frac{\delta^2}{2\kappa}\right)\text{.}
    \end{align}
    If \(\delta > \rho_N\), the right-hand side is smaller than \(\zeta{}\); therefore \(\bm{x} \notin R_N\), which proves \cref{eq:fixed_bandwidth_halo}.
\end{proof}

\begin{remark}[Two fixed-parameter regimes]\label{rem:TwoRegime}
    The first part establishes eventual interior coverage for Lebesgue-almost every point, consistent with the sparse-sample under-approximation observed empirically.
    The outer radius in \cref{eq:fixed_bandwidth_halo}, however, grows like \(\sqrt{\ln N}\), so it does not establish two-sided set convergence at fixed \(\kappa{}\) and \(\lambda{}\).
    Dense noisy-OR superposition can instead expand the boundary unless bandwidths shrink with \(N\).
    The OOD adjustment guarantees exclusion only at the finite OOD samples to which it is applied; it does not change this global fixed-parameter conclusion.
\end{remark}

\FloatBarrier{}

\subsection{Calibrated Kernel Parameters}\label{subsec:AsympSimCalibrated}
For the following calibrated result, let \(R^* \subset \mathbb{R}^n\) be compact with positive finite volume \(V\) and a Lipschitz boundary, and let \(v_n\) denote the volume of the \(n\)-dimensional unit ball.
The Lipschitz boundary implies an interior volume condition: there exist \(r_0 > 0\) and \(0 < b \leq 1\) such that \(\mathcal{L}^n(B(\bm{x}, r) \cap R^*) \geq b \cdot v_n \cdot r^n\) for all \(\bm{x} \in R^*\) and \(r \leq r_0\).
For the first \(N\) anchors, let \(\tilde{d}_N\) denote the median full-space nearest-neighbor distance.
Under the calibrated parameters of \cref{eq:sigma_calibrated}, \(\kappa_N = (s_N \tilde{d}_N)^2\) and \(\lambda_N = \lambda_{\mathrm{rel}} \kappa_N\) shrink as \(N\) grows, while \(\eta_N = \gamma/\tilde{d}_N\) grows.
The following lemmas establish the rates needed for two-sided convergence in measure.

\begin{lemma}[Scaling of the median nearest-neighbor distance]\label{lem:MedianNNScaling}
    There exist constants \(0 < c_1 \leq c_2 < \infty{}\), depending only on \(n\) and \(R^*\), such that almost surely \(c_1 N^{-1/n} \leq \tilde{d}_N \leq c_2 N^{-1/n}\) for all sufficiently large \(N\).
\end{lemma}

\begin{proof}
    Let \(F_N(t)\) denote the fraction of anchor points whose nearest-neighbor distance is at most \(t N^{-1/n}\).
    Conditional on \(\bm{x}_i = \bm{x}\), the remaining \(N - 1\) points avoid \(B(\bm{x}, t N^{-1/n})\) with probability \(\left(1 - \mathcal{L}^n(B(\bm{x}, t N^{-1/n}) \cap R^*)/V\right)^{N - 1}\).
    The interior volume condition and the unit-ball upper bound therefore give
    \begin{align}
        1 - \exp\left(-\frac{b \cdot v_n \cdot t^n}{V}\right) + o(1) \leq \mathbb{E} F_N(t) \leq 1 - \exp\left(-\frac{v_n \cdot t^n}{V}\right) + o(1)\text{.}
    \end{align}
    Thus, \(t_1\) can be chosen small enough that \(\limsup_N \mathbb{E} F_N(t_1) \leq 0.25\), and \(t_2\) can be chosen large enough that \(\liminf_N \mathbb{E} F_N(t_2) \geq 0.75\).
    In fixed dimension, a point can be the nearest neighbor of at most \(\tau_n\) other points, where \(\tau_n\) depends only on \(n\)~\cite{Eppstein1997}.
    Replacing one anchor can therefore change \(F_N(t)\) by at most \((2\tau_n + 1)/N\).
    McDiarmid's bounded-differences inequality~\cite{McDiarmid1989} gives
    \begin{align}
        \mathbb{P}\left(\left|F_N(t) - \mathbb{E} F_N(t)\right| \geq \frac{1}{8}\right) \leq 2\exp\left(-\frac{N}{32(2\tau_n + 1)^2}\right)\text{,}
    \end{align}
    which is summable in \(N\).
    The Borel--Cantelli lemma~\cite{Durrett2019} implies that almost surely \(F_N(t_1) < 0.5 < F_N(t_2)\) for all sufficiently large \(N\).
    Consequently, \(t_1 N^{-1/n} \leq \tilde{d}_N \leq t_2 N^{-1/n}\); related nearest-neighbor laws are given in prior research~\cite{Penrose2003}.
\end{proof}

\begin{lemma}[Local rates at a fixed interior point]\label{lem:LocalRates}
    Fix \(\bm{x} \in \interior(R^*)\), let \(\bm{x}_{(1),N}\) and \(\bm{x}_{(2),N}\) be its nearest and second-nearest anchors, and let \(r_{(2),N}(\bm{x}) := \|\bm{x} - \bm{x}_{(2),N}\|_2\).
    There exists \(C > 0\) such that almost surely \(r_{(2),N}(\bm{x}) \leq (C \ln\ln N/N)^{1/n}\) for all sufficiently large \(N\).
\end{lemma}

\begin{proof}
    The distance \(r_{(2),N}(\bm{x})\) is non-increasing in \(N\).
    Let \(N_k = 2^k\) and \(r_k = (a \ln k/N_k)^{1/n}\), with \(a > 0\) to be chosen.
    For sufficiently large \(k\), \(B(\bm{x}, r_k) \subseteq R^*\), so \(q_k := \mathcal{L}^n(B(\bm{x}, r_k))/V = v_n \cdot a \ln k/(V \cdot N_k)\).
    The probability that this ball contains at most one of the \(N_k\) anchors satisfies
    \begin{align}
        (1 - q_k)^{N_k} + N_k \cdot q_k (1 - q_k)^{N_k - 1} \leq (1 + N_k \cdot q_k) \exp(-(N_k - 1) q_k) \leq C_a(\ln k) k^{-v_n \cdot a/V}\text{.}
    \end{align}
    This bound is summable in \(k\) when \(v_n \cdot a/V \geq 3\).
    The Borel--Cantelli lemma~\cite{Durrett2019} gives \(r_{(2),N_k}(\bm{x}) \leq r_k\) almost surely for all sufficiently large \(k\).
    For \(N_k \leq N < N_{k + 1}\), monotonicity gives \(r_{(2),N}(\bm{x}) \leq (2a \ln k/N)^{1/n}\), and \(\ln k \leq \ln\ln N + 1\) proves the claim.
\end{proof}

For sets \(E\) and \(F\), \(E \mathbin{\triangle} F := (E \setminus F) \cup (F \setminus E)\) denotes their symmetric difference.

The asymptotic results below concern the retained anchor sequence under continuous sampling.
Under their assumptions, exact duplicate observations occur with probability zero, so exact-record de-duplication is almost surely the identity.
Resolution-cell merging for quantized acquisition is a finite-sample preprocessing operation and is not covered by the asymptotic statement.

\begin{theorem}[Asymptotic ODD Similarity, calibrated parameters]\label{thm:AsympODDSimCalibrated}
    Let \(n \geq 2\), let \(\bm{x}_1, \bm{x}_2, \ldots{}\) be independent and uniformly distributed on \(R^*\), and calibrate the parameters as in \cref{eq:sigma_calibrated} with \(s_N = \sqrt{\ln N}\), \(\gamma > 0\), and \(\lambda_{\mathrm{rel}} \in (0, 1)\).
    For every \(\zeta \in (0, 1)\) and \(N \geq 2\), let \(\alpha_N\) be the global affinity constructed from the first \(N\) anchors and let \(R_N := \{\bm{x} : \alpha_N(\bm{x}) \geq \zeta{}\}{}\) be its unadjusted membership region.
    Then, almost surely, Lebesgue-almost every \(\bm{x} \in \interior(R^*)\) belongs to \(R_N\) for all sufficiently large \(N\), every fixed \(\bm{x}\) with \(d(\bm{x}, R^*) > 0\) lies outside \(R_N\) for all sufficiently large \(N\), and
    \begin{align}
        \mathcal{L}^n(R_N \mathbin{\triangle} R^*) \to 0\text{.}
    \end{align}
    Thus, for a shared interpretation function and noise space, the represented ODD approaches the similarity criterion of \cref{def:ODDSimilarity} in the symmetric-difference measure.
\end{theorem}

\begin{proof}
    \noindent\emph{Interior points.}
    Fix \(\bm{x} \in \interior(R^*)\).
    By \cref{lem:LocalRates}, almost surely \(\|\bm{x} - \bm{x}_{(1),N}\|_2 \leq r_{(2),N}(\bm{x}) \leq (C \ln\ln N/N)^{1/n}\) eventually.
    The nearest-neighbor distance of \(\bm{x}_{(1),N}\) satisfies \(d_{(1),N}^* \leq \|\bm{x}_{(1),N} - \bm{x}_{(2),N}\|_2 \leq 2(C \ln\ln N / N)^{1/n}\).
    For either nearest-neighbor mode, the coordinatewise distance used for each diagonal entry is at most \(d_{(1),N}^*\).
    By \cref{lem:MedianNNScaling}, \(\eta_N \cdot d_{(1),N}^* \leq t_N := (2 \gamma C^{1/n} / c_1)(\ln\ln N)^{1/n}\).
    Every diagonal entry of \(\bm{\Sigma}_{(1),N}\) therefore satisfies
    \begin{align}
        \sigma_{kk}^{(1,N)} \geq (1 - \lambda_{\mathrm{rel}}) c_1^2 s_N^2 N^{-2/n} \exp(-t_N)\text{.}
    \end{align}
    It follows that
    \begin{align}
        \alpha_{(1),N}(\bm{x}) \geq \exp\left(-\frac{C^{2/n} (\ln\ln N)^{2/n} \exp(t_N)}{2(1 - \lambda_{\mathrm{rel}}) c_1^2 \ln N}\right) \to 1\text{.}
    \end{align}
    Indeed, the logarithm of the exponent magnitude is \((2/n) \ln\ln\ln N + t_N - \ln\ln N + O(1)\), which tends to \(-\infty{}\) because \(n \geq 2\).
    Hence \(\alpha_N(\bm{x}) \geq \alpha_{(1),N}(\bm{x}) \geq \zeta{}\) eventually.

    \medskip
    \noindent\emph{Exterior points.}
    Let \(\delta = d(\bm{x}, R^*) > 0\).
    Every kernel satisfies \(\sigma_{kk}^{(i)} \leq \kappa_N\), so the noisy-OR union bound and \cref{lem:MedianNNScaling} give
    \begin{align}
        \alpha_N(\bm{x}) \leq N \exp\left(-\frac{\delta^2}{2\kappa_N}\right) \leq N \exp\left(-\frac{\delta^2 N^{2/n}}{2c_2^2 \ln N}\right) \to 0\text{.}
    \end{align}
    Moreover, the represented region is globally contained in a shrinking dilation because
    \begin{align}
        \rho_N^{\mathrm{cal}} := \sqrt{2\kappa_N \ln\left(\frac{N}{\zeta}\right)} = s_N \tilde{d}_N \sqrt{2\ln\left(\frac{N}{\zeta}\right)} \to 0\text{.}
    \end{align}

    \medskip
    \noindent\emph{Convergence in measure.}
    The preceding steps give pointwise indicator convergence for every fixed point outside \(\partial R^*\), almost surely.
    Fubini's theorem~\cite{Evans2025} makes this conclusion simultaneous for Lebesgue-almost every point, and the boundary is Lebesgue-null by the Lipschitz assumption.
    The shrinking outer containment supplies a common compact superset, so the dominated convergence theorem~\cite{Evans2025} yields \(\mathcal{L}^n(R_N \mathbin{\triangle} R^*) \to 0\).
\end{proof}

\FloatBarrier{}

\section{Justification of Diagonal Sigma-Matrices}\label{sec:JustDiagSigma}

\subsection{Theoretical Justification}
The deployed covariance matrices are diagonal but anisotropic: each coordinate receives its own local scale according to \cref{eq:sigma_kk_mod}.
This choice represents axis-aligned changes in local sampling density while retaining linear storage and evaluation cost in the input dimension.
It does not, however, represent rotated anisotropy or correlations between coordinates.

A diagonal approximation to a full local covariance is statistically justified only when the off-diagonal population correlations are negligible, i.e., when the coordinate components of nearby anchor points are approximately uncorrelated in the fixed working coordinate system.
Local uniformity by itself is insufficient: a locally elongated distribution may have persistent cross-coordinate correlation, even with infinitely many samples.
Consequently, the diagonal form is an explicit accuracy--efficiency trade-off whose discrepancy from a corresponding full-covariance structure is quantified below.
This trade-off concerns finite-sample accuracy, not consistency: \cref{thm:AsympODDSimCalibrated} assumes a Lipschitz boundary but no independence between coordinates, so the calibrated diagonal construction also converges in volume for coupled parameters.

\FloatBarrier{}

\subsection{Approximation Error Bound}

Let \(\bm{\Sigma}_i^{\mathrm{full}}\) denote a symmetric positive-definite full covariance and let \(\bm{\Sigma}_i^{\mathrm{diag}} = \diag(\bm{\Sigma}_i^{\mathrm{full}})\) be its diagonal projection.
This comparison isolates covariance structure; \(\bm{\Sigma}_i^{\mathrm{full}}\) is not an estimator used by the deployed method.
For \(\bm{v} = \bm{x} - \bm{x}_i\), define
\begin{align}
    q_i^{\mathrm{full}}(\bm{x})
    = \bm{v}^\top \left(\bm{\Sigma}_i^{\mathrm{full}}\right)^{-1} \bm{v},
    \qquad
    q_i^{\mathrm{diag}}(\bm{x})
    = \bm{v}^\top \left(\bm{\Sigma}_i^{\mathrm{diag}}\right)^{-1} \bm{v}\text{.}
\end{align}

The resolvent identity gives the exact relation
\begin{align}
    \left(\bm{\Sigma}_i^{\mathrm{full}}\right)^{-1}
    -\left(\bm{\Sigma}_i^{\mathrm{diag}}\right)^{-1}
    =
    \left(\bm{\Sigma}_i^{\mathrm{full}}\right)^{-1}
    \left(\bm{\Sigma}_i^{\mathrm{diag}} - \bm{\Sigma}_i^{\mathrm{full}}\right)
    \left(\bm{\Sigma}_i^{\mathrm{diag}}\right)^{-1}\text{.}
\end{align}
The mean value theorem applied to \(t \mapsto \exp(-t/2)\), followed by the spectral norm bound for a quadratic form, therefore yields
\begin{align}
     & \left|\alpha_i^{\mathrm{full}}(\bm{x})
    -\alpha_i^{\mathrm{diag}}(\bm{x})\right|
    \leq \frac{1}{2}\left\|\left(\bm{\Sigma}_i^{\mathrm{full}}\right)^{-1}\right\|_2
    \left\|\bm{\Sigma}_i^{\mathrm{diag}} - \bm{\Sigma}_i^{\mathrm{full}}\right\|_2
    \nonumber                                 \\
     & \quad
    \left\|\left(\bm{\Sigma}_i^{\mathrm{diag}}\right)^{-1}\right\|_2
    \|\bm{v}\|_2^2
    \max\left\{\alpha_i^{\mathrm{full}}(\bm{x}),
    \alpha_i^{\mathrm{diag}}(\bm{x})\right\}\text{.}
    \label{eq:diag_full_exact_bound}
\end{align}
Thus, the local discrepancy is controlled by the omitted off-diagonal structure and by the conditioning of both covariance matrices.
It vanishes for an axis-aligned covariance, but it need not vanish merely because more locally correlated samples are observed.
For the global noisy-OR affinity, a telescoping product bound gives
\begin{align}
    \left|\alpha^{\mathrm{full}}(\bm{x})
    -\alpha^{\mathrm{diag}}(\bm{x})\right|
    \leq \sum_{i = 1}^{N}
    \left|\alpha_i^{\mathrm{full}}(\bm{x})
    -\alpha_i^{\mathrm{diag}}(\bm{x})\right|\text{,}
\end{align}
so \cref{eq:diag_full_exact_bound} can be summed over the kernels that contribute materially at \(\bm{x}\).
The covariance probe in \cref{fig:CovarianceStructure} uses the rotated-band region \(\{(x_1, x_2) \in [-5, 5]^2 : \lvert x_2 - \rho x_1\rvert \leq 0.6\}{}\), where \(\rho{}\) controls the rotational coupling.

\begin{figure}[htb]
    \begin{subfigure}[t]{0.475\linewidth}
        \centering
        \resizebox{\linewidth}{!}{%
            \begin{tikzpicture}
    \begin{axis}[
        autosafe plot,
        xmin=0,
        xmax=3,
        ymin=0,
        ymax=0.3,
        ylabel={IoU at \(\zeta = 0.5\)},
        xlabel={Coupling \(\rho{}\)},
        xtick={0,1,2,3},
        legend pos=north east,
    ]
        \addplot[color=DLRDarkerYellow,mark=triangle*] table[x=rho,y=diag]{graphics/data/benchmark/covariance_n100.dat};
        \addlegendentry{Diagonal}
        \addplot[color=DLRDarkerGreen,mark=square*] table[x=rho,y=full]{graphics/data/benchmark/covariance_n100.dat};
        \addlegendentry{Full}
        \addplot[color=DLRDarkGrey,densely dotted,mark=o,mark options={fill=white}] table[x=rho,y=iso]{graphics/data/benchmark/covariance_n100.dat};
        \addlegendentry{Isotropic}
    \end{axis}
    \autosafepairedplotbounds{40pt}{11pt}{34pt}{7pt}
\end{tikzpicture}%
        }%
        \caption{Controlled covariance-structure comparison at \(N = 100\). The case \(\rho = 0\) is axis-aligned, whereas \(\rho = 1\) and \(\rho = 3\) rotate the narrow ODD relative to the working coordinates.}\label{fig:CovarianceStructure-Probe}
    \end{subfigure}
    \hfill
    \begin{subfigure}[t]{0.475\linewidth}
        \centering
        \resizebox{\linewidth}{!}{%
            \begin{tikzpicture}
    \begin{axis}[
        autosafe plot,
        xmode=log,
        log basis x=10,
        xtickten={2,3},
        minor xtick={200,300,400,500,600,700,800,900},
        xmin=100,
        xmax=1000,
        ymin=0,
        ymax=1,
        xlabel={Anchor count \(N\)},
        ylabel={IoU at \(\zeta = 0.5\)},
        legend pos=south east,
    ]
        \addplot[color=DLRDarkerBlue,mark=*] table[x=n,y=rho0]{graphics/data/benchmark/deployed_covariance.dat};
        \addlegendentry{\(\rho = 0\)}
        \addplot[color=DLRDarkerGreen,mark=square*] table[x=n,y=rho1]{graphics/data/benchmark/deployed_covariance.dat};
        \addlegendentry{\(\rho = 1\)}
        \addplot[color=DLRDarkerYellow,mark=triangle*] table[x=n,y=rho3]{graphics/data/benchmark/deployed_covariance.dat};
        \addlegendentry{\(\rho = 3\)}
    \end{axis}
    \autosafepairedplotbounds{40pt}{11pt}{34pt}{7pt}
\end{tikzpicture}%
        }%
        \caption{IoU of the deployed diagonal construction as the anchor count increases. All three curves improve with additional anchors, reaching \num{0.916}, \num{0.808}, and \num{0.744} at \(N = \num{1000}\).}\label{fig:CovarianceStructure-Deployed}
    \end{subfigure}
    \caption{Quantitative covariance-structure probe under increasing rotational coupling. At \(N = 100\), full and diagonal covariance produce nearly identical IoU for the axis-aligned case, while full covariance raises the IoU from \num{0.187} to \num{0.261} at \(\rho = 1\) and from \num{0.165} to \num{0.244} at \(\rho = 3\). The measured gap quantifies the expected loss when the local-independence assumption is violated, while the increasing curves of the deployed construction show that greater anchor density still improves the represented region at every tested coupling strength.}\label{fig:CovarianceStructure}
\end{figure}

\FloatBarrier{}

\section{Implementation Details}\label{sec:ImplDetails}
\subsection{Numerical Evaluation of the Affinity}\label{subsec:NumericalAffinity}
The product in \cref{eq:Superposition} is exact in real arithmetic, but direct finite-precision evaluation can absorb small local contributions, underflow the survival product, and round the final subtraction to an affinity of exactly one.
The precise breakpoints depend on the chosen arithmetic; the failure is therefore not specific to one floating-point format.
For the full VCAS run with \(N = \num{622110}\) anchors (cf.\ \cref{subsec:ValidationviaUseCase}), the elementary bound \(\alpha(\bm{x}) \geq 1 - \exp(-\sum_i \alpha_i(\bm{x}))\) leads to exponentially small survival probabilities where the accumulated local affinity is large; in direct finite-precision evaluation, \(\alpha(\bm{x}) = 1\) for \qty{26.3}{\percent} of the evaluation samples, causing ordinary affinity thresholds to lose distinctions within that saturated subset.

Therefore, the affinity is evaluated in log-space.
Defining the log-survival function
\begin{align}
    S(\bm{x}) = \ln\left(1 - \alpha(\bm{x})\right) = \sum_i \ln\left(1 - \alpha_i(\bm{x})\right) \label{eq:LogSurvival}
\end{align}
turns the product into a sum of same-signed terms.
Each term is computed with a stable branchwise scheme~\cite{Maechler2015}, and the accumulated rounding error is bounded by standard summation results~\cite{Higham2002}.
Since \(t \mapsto \ln(1 - t)\) is strictly decreasing, the membership rule transforms exactly:
\begin{align}
    \alpha(\bm{x}) \geq \zeta \iff S(\bm{x}) \leq \ln(1 - \zeta)\text{.} \label{eq:LogThreshold}
\end{align}
All reported results use \cref{eq:LogThreshold}.

\FloatBarrier{}

\subsection{Software Interface}
The Python-based tool provides a sophisticated interface for various applications, including Monte Carlo--based sampling for validation (cf.\ \cref{subsec:MonteCarloValidation}) and the provision of real-world data in different formats (cf.\ \cref{subsec:ValidationviaUseCase}).
The tool can parse data from a wide range of sources, including CSV files, NumPy-like arrays, dataframes, and, most importantly, the JSON-based ASAM OpenLABEL format~\cite{VFSAM2021a}.
The latter is an annotation format commonly used in the automotive domain, designed to facilitate the labeling and tracking of objects across multiple scenes.
At ingestion, the ID records that define the ODD geometry are first de-duplicated using documented acquisition-resolution metadata while preserving merged source identifiers and visit counts.
The retained records are then converted into an internal kernel-based representation that enables efficient querying of the affinity value for one or more sample points.
Thus, it can be integrated into current development methodologies and used as an online tool to determine whether new, unseen samples fall within the ODD, thereby enhancing the safety of already deployed AI-based systems.
This is achieved using a set of libraries, such as NumPy~\cite{Harris2020} and SciPy~\cite{Virtanen2020} for efficient data handling, polytope~\cite{Filippidis2016} for representing ODD polytopes (cf.\ \cref{sec:MathODD}), and Faiss~\cite{Johnson2021, Douze2026} for fast nearest-neighbor searches.

\clearpage{}

\section{The autoSAFE Algorithm}
\begin{algorithm}[htb]
    \caption{Automated Kernel-Based ODD Derivation for RBF Kernels}\label{alg:autoSAFE}
    \begin{algorithmic}[1]
        \STATE \textbf{Input:}
        \STATE \quad ID record collection \(\widetilde{\mathcal{D}}_{\mathrm{ID}}\) and OOD sample set \(\mathcal{D}_{\mathrm{OOD}}\)
        \STATE \quad Documented acquisition-grid metadata \((\bm{q}, \bm{o})\)
        \STATE \quad Calibration constants \(\gamma{}\) and \(\lambda_{\mathrm{rel}}\), scheduling rule \(s_N = \sqrt{\ln N}\)
        \STATE \quad OOD affinity threshold \(\xi{}\)
        \STATE \quad Covariance shrink factor \(c \in (0, 1)\)
        \STATE \quad False-exclusion budget \(\varepsilon \in (0, 1)\) and calibration-set size \(N_{\mathrm{cal}} \geq 1/\varepsilon - 1\)
        \STATE \quad Optional anchor count \(N\)
        \STATE \quad Documented random seed for the calibration split and the optional anchor subsampling
        \STATE \textbf{Calibration Split:}
        \STATE Reserve \(N_{\mathrm{cal}}\) held-out records \(\mathcal{D}_{\mathrm{cal}}\), selected uniformly at random from \(\widetilde{\mathcal{D}}_{\mathrm{ID}}\), and remove them from \(\widetilde{\mathcal{D}}_{\mathrm{ID}}\)
        \STATE \textbf{Deterministic ID Anchor De-duplication:}
        \STATE Partition \(\widetilde{\mathcal{D}}_{\mathrm{ID}}\) into acquisition-resolution cells defined by \((\bm{q}, \bm{o})\)
        \FOR{each occupied cell \(C\)}
        \STATE Sort the records in \(C\) lexicographically and compute their mean \(\bar{\bm{x}}_C\) in that order
        \STATE Retain the lexicographically first minimizer of \(\sum_{k=1}^{n} ((x_k - \bar{x}_{Ck}) / q_k)^2\) over \(\bm{x} \in C\)
        \STATE Store the merged source identifiers and record count as provenance metadata
        \ENDFOR
        \STATE Index the retained ID representatives and OOD samples in lexicographic order
        \STATE Set \(\mathcal{D}_{\mathrm{ID}}\) to the canonically indexed ID representatives
        \STATE Set anchor points \(\mathcal{A} \leftarrow \mathcal{D}_{\mathrm{ID}}\), or, if \(N\) is given, select \(N\) of them uniformly at random without replacement, retaining their canonical order
        \STATE \textbf{Kernel Parameter Estimation:}
        \STATE Compute the full-space nearest-neighbor distances and their median \(\tilde{d}\)
        \STATE Set \(\eta \leftarrow \gamma/\tilde{d}\), \(\kappa \leftarrow (s_N \tilde{d})^2\), and \(\lambda \leftarrow \lambda_{\mathrm{rel}} \kappa{}\)
        \FOR{each anchor point \(\bm{x}_i \in \mathcal{A}\)}
        \STATE Compute the selected global or per-dimension exact nearest-neighbor distances \(d_{ik}^*\), \(k = 1, \dots, n\)
        \COMMENT{lowest canonical anchor index resolves global-mode ties}
        \FOR{each dimension \(k\)}
        \STATE Set \(\sigma_{kk}^{(i)} \leftarrow (\kappa - \lambda) \exp(-\eta d_{ik}^*) + \lambda{}\)
        \ENDFOR
        \STATE Define kernel \(\alpha_i(\bm{x})\) using \(\bm{\Sigma}_i\)
        \ENDFOR
        \STATE \textbf{OOD Consistency Check (cf.\ \cref{def:OODConsCons}):}
        \COMMENT{requires \(\mathcal{A} \cap \mathcal{D}_\mathrm{OOD} = \emptyset{}\)}
        \WHILE{\(\exists \bm{x}_\mathrm{OOD} \in \mathcal{D}_\mathrm{OOD}\) s.t.\ \(\alpha(\bm{x}_\mathrm{OOD}) > \xi{}\)}
        \STATE \(\bm{x}^* \leftarrow \argmax_{\bm{x} \in \mathcal{D}_\mathrm{OOD}} \alpha(\bm{x})\)
        \COMMENT{globally most-violated OOD point}
        \STATE \(i^* \leftarrow \argmax_{i} \alpha_i(\bm{x}^*)\)
        \COMMENT{dominant kernel}
        \STATE Adjust \(\bm{\Sigma}_{i^*} \leftarrow c \cdot \bm{\Sigma}_{i^*}\text{,}\quad c \in (0, 1)\)
        \ENDWHILE
        \STATE \textbf{Membership Threshold Calibration (cf.\ \cref{subsec:ConformalThreshold}):}
        \STATE Compute \(Q(\bm{x}) = -\ln(1 - \alpha(\bm{x}))\) for all \(\bm{x} \in \mathcal{D}_{\mathrm{cal}}\) and sort them in ascending order
        \STATE Set \(k \leftarrow \lfloor \varepsilon(N_{\mathrm{cal}} + 1)\rfloor{}\), \(\hat{t} \leftarrow Q_{(k)}\), and \(\hat{\zeta} \leftarrow 1 - \exp(-\hat{t})\)
        \STATE Verify \(\xi < \hat{\zeta}\)
        \STATE \textbf{Output:} Deterministic kernel-based ODD affinity function \(\alpha(\bm{x})\) and membership threshold \(\hat{\zeta}\)
    \end{algorithmic}
\end{algorithm}

\FloatBarrier{}

\section{Computational Scalability}\label{sec:ComputationalScalability}
The pseudocode in \cref{alg:autoSAFE} has total construction cost \(O(N^2 n + N_{\mathrm{adj}}MNn + N_{\mathrm{cal}}Nn)\) and model-storage cost \(O(Nn)\).
Here, \(N\) is the number of anchors, \(n\) is the parameter-space dimension, \(M\) is the number of OOD samples, and \(N_{\mathrm{adj}}\) is the finite number of covariance-adjustment steps bounded in \cref{prop:OODTermination}.
The term \(N^2 n\) arises because the direct nearest-neighbor construction compares every pair of anchors in \(n\) dimensions.
Each OOD adjustment evaluates all \(N\) kernels at all \(M\) OOD samples, which accounts for \(MNn\) operations per step and \(N_{\mathrm{adj}}MNn\) in total.
Calibrating \(\hat{\zeta}\) evaluates all \(N\) kernels at the \(N_{\mathrm{cal}}\) held-out records, which accounts for \(N_{\mathrm{cal}}Nn\) operations.
After construction, an exact affinity query evaluates all local affinities and the noisy-OR aggregation in \(O(Nn)\) time.

An exact nearest-neighbor index with \(O(n \log N)\) query cost in fixed dimension can replace the pairwise search, reducing the first term to \(O(Nn \log N)\) under the index's stated assumptions without changing the mathematical construction.
A separate approximation retains only the \(K_\mathrm{near}\) kernels whose anchors are nearest in normalized Euclidean distance during affinity evaluation, yielding \(O(n \log N + K_\mathrm{near} n)\) time per query when paired with such an index.
Because the RBF contributions are positive, truncation always underestimates the exact noisy-OR affinity, but the omitted tail and the resulting membership decisions depend on \(K_\mathrm{near}\), the anchor density, and the kernel widths.
The quantitative experiment in \cref{fig:TruncationScalability} therefore reports affinity error, decision changes, runtime, and host-side working-set memory rather than treating truncation as exact.

For \(K_\mathrm{near} = 512\), the maximum absolute affinity error increases from \num{0.018} at \(N = \num{1000}\) to \num{0.434} at \(N = \num{6e5}\); at the largest anchor count, the mean and 99th-percentile errors are \num{0.0303} and \num{0.257}.
For \(K_\mathrm{near} = 128\) at \(N = \num{6e5}\), the corresponding maximum, mean, and 99th-percentile errors are \num{0.696}, \num{0.0687}, and \num{0.503}.
At this anchor count, the anchor and inverse-diagonal model arrays shared by both evaluators occupy \qty{48}{\mega\byte}, while the Python-traced temporary working set of the truncated implementation peaks at \qty{184}{\mega\byte} for \(K_\mathrm{near} = 128\) and \qty{737}{\mega\byte} for \(K_\mathrm{near} = 512\).
These working-set measurements exclude JAX/XLA backend allocations used by the exact evaluator and therefore do not establish a measured memory advantage over exact evaluation.
Truncation is therefore studied only as an optional acceleration; the implementation and all other reported results use exact evaluation.

\begin{figure}[htb]
    \begin{subfigure}[t]{0.475\linewidth}
        \centering
        \resizebox{\linewidth}{!}{%
            \begin{tikzpicture}
    \begin{axis}[
        autosafe plot,
        xmode=log,
        log basis x=2,
        log ticks with fixed point,
        xtickten={3,5,7,9},
        xmin=8,
        xmax=512,
        xlabel={Retained kernels \(K_\mathrm{near}\)},
        ylabel={Changed decisions at \(\zeta = 0.5\) in \unit{\percent}},
        ymin=0,
        ymax=35,
        legend pos=north east,
    ]
        \addplot[color=DLRDarkerYellow,mark=triangle*]
            table[x=k,y expr={100*\thisrow{flip_zeta05}},restrict expr to domain={\thisrow{n}}{9999:10001}]{graphics/data/benchmark/truncation_error.dat};
        \addlegendentry{\(N = 10^4\)}
        \addplot[color=DLRDarkerGreen,mark=square*]
            table[x=k,y expr={100*\thisrow{flip_zeta05}},restrict expr to domain={\thisrow{n}}{99999:100001}]{graphics/data/benchmark/truncation_error.dat};
        \addlegendentry{\(N = 10^5\)}
        \addplot[color=DLRDarkerBlue,mark=*]
            table[x=k,y expr={100*\thisrow{flip_zeta05}},restrict expr to domain={\thisrow{n}}{599999:600001}]{graphics/data/benchmark/truncation_error.dat};
        \addlegendentry{\(N = 6 \cdot 10^5\)}
    \end{axis}
    \autosafepairedplotbounds{38pt}{11pt}{34pt}{8pt}
\end{tikzpicture}%
        }%
        \caption{Fraction of changed membership decisions at \(\zeta = 0.5\) as the number of retained kernels increases for three anchor counts.}\label{fig:TruncationScalability-Accuracy}
    \end{subfigure}
    \hfill
    \begin{subfigure}[t]{0.475\linewidth}
        \centering
        \resizebox{\linewidth}{!}{%
            \begin{tikzpicture}
    \begin{axis}[
        autosafe plot,
        xmode=log,
        log basis x=10,
        xtickten={4,5},
        minor xtick={20000,30000,40000,50000,60000,70000,80000,90000,200000,300000,400000,500000,600000},
        xmin=10000,
        xmax=600000,
        ymin=0,
        ymax=5,
        xlabel={Anchor count \(N\)},
        ylabel={Runtime for \(10^4\) queries in \unit{\second}},
        ytick={0,1,2,3,4,5},
        yticklabels={0,1,2,3,4,5},
        legend pos=north west,
    ]
        \addplot[color=DLRDarkerBlue,mark=*] table[x=n,y=exact]{graphics/data/benchmark/truncation_latency.dat};
        \addlegendentry{Exact}
        \addplot[color=DLRDarkerGreen,mark=square*] table[x=n,y=k128]{graphics/data/benchmark/truncation_latency.dat};
        \addlegendentry{\(K_\mathrm{near} = 128\)}
        \addplot[color=DLRDarkerYellow,mark=triangle*] table[x=n,y=k512]{graphics/data/benchmark/truncation_latency.dat};
        \addlegendentry{\(K_\mathrm{near} = 512\)}
    \end{axis}
    \autosafepairedplotbounds{38pt}{11pt}{34pt}{8pt}
\end{tikzpicture}%
        }%
        \caption{Wall-clock runtime for \(\num{10000}\) queries under exact evaluation and two truncation levels as the number of anchors increases.}\label{fig:TruncationScalability-Runtime}
    \end{subfigure}
    \caption{Accuracy and runtime of \(K_\mathrm{near}\)-kernel truncation on the five-dimensional constrained ODD with \(\num{10000}\) queries. At \(N = \num{6e5}\), \(K_\mathrm{near} = 512\) reduces runtime from \(\qty{3.17}{\second}\) to \(\qty{1.52}{\second}\) with \(\qty{3.22}{\percent}\) changed decisions at \(\zeta = 0.5\), whereas \(K_\mathrm{near} = 128\) takes \(\qty{0.63}{\second}\) with \(\qty{7.11}{\percent}\) changed decisions. The measured crossover is approximately \(\num{2.3e5}\) anchors for \(K_\mathrm{near} = 512\); below it, the fixed costs of nearest-neighbor lookup and selected-kernel gathering plausibly outweigh the kernel evaluations avoided by truncation.}\label{fig:TruncationScalability}
\end{figure}

\FloatBarrier{}

\section{Description of Aviation Use Cases}\label{sec:ACASX}
In aviation, one of the most important tasks of pilots during en route travel is avoiding collisions with other aircraft.
The next generation of collision avoidance systems is expected to improve performance significantly; however, it has become apparent that the proposed system cannot run on current state-of-the-art avionics hardware because its memory requirements are too high~\cite{ED-275, Damour2021}.
Preliminary minimal implementations split the system into a horizontal and a vertical component, called the Horizontal Collision Avoidance System (HCAS) and the Vertical Collision Avoidance System (VCAS), respectively, using neural networks for policy compression~\cite{Julian2016, Julian2019, Julian2019a}.
Thus, a stringent safety argumentation will be required for an AI-based collision avoidance system~\cite{Christensen2024}.
Given that the standards clearly define the operational conditions, an appropriate ODD can be derived and later compared with the ODD derived by autoSAFE\@.
Therefore, this use case is a prime example for further study in this work.
However, in line with previous research~\cite{Julian2019}, only a subset of the full operational conditions will be examined.

\subsection{Horizontal Collision Avoidance System}
\begin{figure}[htb]
    \centering
    \includegraphics[scale=0.85]{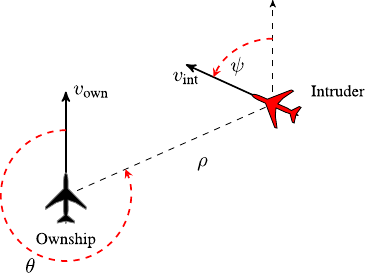}
    \caption{Geometry of the horizontal collision avoidance scenario for HCAS, from~\cite{Julian2019}. The black ownship is trying to avoid the (malicious) red intruder by diverting either to the left or to the right.}\label{fig:HCAS}
\end{figure}

For HCAS, the decision depends on the range and bearing angle to the intruder, the relative heading angle, the ownship's and intruder's speed, the time to CPA, and the advisory issued in the previous time step.
A geometric overview of this problem is given in \cref{fig:HCAS}.
Furthermore, the parameter ranges are shown in \cref{tab:HCAS_ODD}.
\begin{table}[htb]
    \caption{Parameter ranges defining the taxonomy \(X^*\) in the reported HCAS evaluation. The ranges are a subset of the original HCAS implementation~\cite{Julian2019}.}\label{tab:HCAS_ODD}
    \centering
    \begin{tabular}{lll}
        \toprule
        Variable           & Description       & Range                                                            \\
        \midrule
        \(\rho{}\)         & Range to intruder & \([\qty{0}{\foot}, \qty{50000}{\foot}]\)                       \\
        \(\theta{}\)       & Bearing angle     & \([\ang{-180}, \ang{180}]\)                                      \\
        \(\psi{}\)         & Relative heading  & \([\ang{-180}, \ang{180}]\)                                      \\
        \(v_\mathrm{own}\) & Ownship speed     & \([\qty{840}{\foot\per\second}, \qty{870}{\foot\per\second}]\) \\
        \(v_\mathrm{int}\) & Intruder speed    & \([\qty{760}{\foot\per\second}, \qty{870}{\foot\per\second}]\) \\
        \(\tau{}\)         & Time to CPA       & \([\qty{0}{\second}, \qty{80}{\second}]\)                        \\
        \(s_\mathrm{adv}\) & Previous advisory & \(\{0, 1, 2, 3, 4\}\)                                            \\
        \bottomrule
    \end{tabular}
\end{table}
\subsubsection{Known HCAS ODD Structure}
Following \cref{def:ODDStructure}, the known HCAS ODD is defined as the structure
\begin{align}
    \mathcal{O}^*_{\mathrm{HCAS}} = \left(X^*, f^{\mathcal{O}^*}, \Omega^{\mathcal{O}^*}\right)\text{,}
\end{align}
where the taxonomy \(X^* \subset \mathbb{R}^7\) is the hyperrectangle
\begin{align}
    X^* ={} & [\qty{0}{\foot}, \qty{50000}{\foot}] \times [\ang{-180}, \ang{180}] \times [\ang{-180}, \ang{180}] \nonumber                                      \\
            & {}\times [\qty{840}{\foot\per\second}, \qty{870}{\foot\per\second}] \times [\qty{760}{\foot\per\second}, \qty{870}{\foot\per\second}] \nonumber \\
            & {}\times [\qty{0}{\second}, \qty{80}{\second}] \times \{0, 1, 2, 3, 4\} \text{,}
\end{align}
spanned by \((\rho, \theta, \psi, v_{\mathrm{own}}, v_{\mathrm{int}}, \tau, s_{\mathrm{adv}})\) as specified in \cref{tab:HCAS_ODD} and following prior research~\cite{Julian2019}.
No additional relational constraints beyond the parameter ranges are imposed for this use case, i.e., \(r = 0\) and \(\mathcal{R}^{\mathcal{O}^*} = X^*\).
The interpretation function \(f^{\mathcal{O}^*}\colon X^* \to Y\) is the deterministic HCAS policy from~\cite{Julian2019}, mapping each state vector \(\bm{x} \in X^*\) to an advisory \(y \in Y = \{0, 1, 2, 3, 4\}\).
Since the policy is deterministic, the noise space is trivial; thus \(\Omega^{\mathcal{O}^*} = \{0\}\).

\subsubsection{Special Variables}
The previous advisory \(s_{\mathrm{adv}} \in \{0, 1, 2, 3, 4\}\) is a discrete categorical variable.
For the kernel computation, its codes are retained as a scalar coordinate with unit spacing; the kernel metric therefore treats numerical code separation as geometric distance.
The bearing angle \(\theta{}\) and the relative heading \(\psi{}\) are periodic; the validation samples and the known-ODD membership respect this periodicity, whereas the kernel metric treats both as ordinary coordinates, so that states on either side of \(\pm\ang{180}\) count as distant.
All seven parameters are retained, giving a seven-dimensional effective parameter space.

\subsection{Vertical Collision Avoidance System}
\begin{figure}[htb]
    \centering
    \includegraphics[scale=0.85]{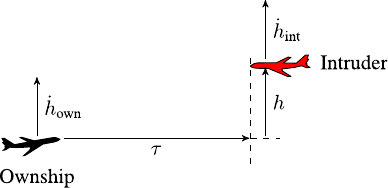}
    \caption{Geometry of the vertical collision avoidance scenario for VCAS, from~\cite{Julian2019}. The black ownship is trying to avoid the (malicious) red intruder by diverting through climbing or descending.}\label{fig:VCAS}
\end{figure}

For VCAS, the decision depends on the relative altitude between the ownship and the intruder, their corresponding vertical rates, the time to CPA, and the advisory issued in the previous time step.
A geometric overview of this problem is given in \cref{fig:VCAS}.
Furthermore, the parameter ranges are shown in \cref{tab:VCAS_ODD}.
\begin{table}[htb]
    \caption{Parameter ranges of the nominal five-dimensional state space in the reported VCAS evaluation. The ranges are a subset of the original VCAS implementation~\cite{Julian2019}.}\label{tab:VCAS_ODD}
    \centering
    \begin{tblr}{colspec={lll}}
        \toprule
        Variable                 & Description          & Range                                                           \\
        \midrule
        \(h\)                    & Relative altitude    & \([\qty{-3000}{\foot}, \qty{3000}{\foot}]\)                   \\
        \(\dot{h}_\mathrm{own}\) & Ownship vert.\ rate  & \([\qty{-3000}{\foot\per\minute}, \qty{3000}{\foot\per\minute}]\) \\
        \(\dot{h}_\mathrm{int}\) & Intruder vert.\ rate & \(\qty{0}{\foot\per\minute}\)                                  \\
        \(\tau{}\)               & Time to CPA          & \([\qty{0}{\second}, \qty{40}{\second}]\)                       \\
        \(s_\mathrm{adv}\)       & Previous advisory    & \(\{0, 1, 2, \dots, 8\}{}\)                                     \\
        \bottomrule
    \end{tblr}
\end{table}

\subsubsection{Known VCAS ODD Structure}
Following \cref{def:ODDStructure}, the known VCAS ODD is defined as the structure
\begin{align}
    \mathcal{O}^*_{\mathrm{VCAS}} = \left(X^*, f^{\mathcal{O}^*}, \Omega^{\mathcal{O}^*}\right)\text{,}
\end{align}
where the taxonomy \(X^* \subset \mathbb{R}^4\) is the product space
\begin{align}
    X^* = [\qty{-3000}{\foot}, \qty{3000}{\foot}]
    \times
    [\qty{-3000}{\foot\per\minute}, \qty{3000}{\foot\per\minute}]
    \times
    [\qty{0}{\second}, \qty{40}{\second}]
    \times
    \{0, 1, 2, \dots, 8\}\text{,}
\end{align}
spanned by \((h, \dot{h}_{\mathrm{own}}, \tau, s_{\mathrm{adv}})\) as specified in \cref{tab:VCAS_ODD} and following prior research~\cite{Julian2019}.
No additional relational constraints beyond the parameter ranges are imposed for this use case, i.e., \(r = 0\) and \(\mathcal{R}^{\mathcal{O}^*} = X^*\).
The interpretation function \(f^{\mathcal{O}^*}\colon X^* \to Y\) is the deterministic VCAS policy from~\cite{Julian2019}, mapping each state vector \(\bm{x} \in X^*\) to an advisory \(y \in Y = \{0, 1, \dots, 8\}{}\).
Since the policy is deterministic, the noise space is trivial; thus \(\Omega^{\mathcal{O}^*} = \{0\}{}\).

\FloatBarrier{}

\subsubsection{Special Variables}
Two parameters require specific treatment.
First, the previous advisory \(s_{\mathrm{adv}} \in \{0, 1, \dots, 8\}{}\) is a discrete categorical variable; it is encoded as a continuous coordinate with unit spacing between adjacent values, which is appropriate because adjacent advisory values reflect operationally similar prior commands.
Second, the intruder vertical rate \(\dot{h}_{\mathrm{int}}\) takes the single value \(\qty{0}{\foot\per\minute}\) across all states in the dataset, yielding zero variance in that dimension.
A zero-variance dimension produces a degenerate covariance matrix entry and contributes no discriminative information to the kernel; it is therefore excluded from the kernel computation and from \(X^*\) above.
This reduces the effective parameter space from the nominal 5D to 4D.

\clearpage{}

\section{EASA Objectives Addressed by autoSAFE}\label{sec:EASAObjectives}
The operational domain (OD) describes the aircraft- or system-level operating conditions, while the ODD applies at the AI/ML constituent level.
The quoted objectives below abbreviate data quality requirements as DQRs.
\Cref{tab:EASAObjectives} relates properties of autoSAFE to selected objectives of the EASA concept paper, Issue 02~\cite{EUASA2024}; the corresponding identifiers in the proposed Issue 03~\cite{EUASA2026} are shown in parentheses.
The mapping identifies potential contributions to an assurance case, not compliance with an objective, and each contribution remains use-case specific.

\begin{longtblr}[
        caption = {Selected objectives from the EASA concept paper to which autoSAFE can contribute. Issue 02 identifiers are followed by their counterparts in proposed Issue 03~\cite{EUASA2024, EUASA2026}.},
        label = {tab:EASAObjectives},
    ]{
        colspec = {Q[l,wd=0.22\linewidth]XX},
        rowhead = 1,
        rowfoot = 0,
    }
    \toprule
    \textbf{Objective} &
    \textbf{Excerpt}   &
    \textbf{Rationale}   \\
    \midrule

    DA-03 (SURK-DA-03)
                       &
    \enquote{The applicant should define the set of parameters pertaining to the AI/ML constituent ODD, and trace them to the corresponding parameters pertaining to the OD when applicable.}
                       &
    The framework represents the ODD taxonomy and its interdependencies through an explicit membership function derived from documented operational records.
    Traceability from anchors to the OD depends on the provenance maintained for those records by the surrounding data-management process.
    \\
    \midrule

    DA-04 (SU-DA-04)
                       &
    \enquote{The applicant should capture the DQRs for all data required for training, testing, and verification of the AI/ML constituent~[\ldots]}
                       &
    The affinity provides a per-sample measure of support relative to the constructed ODD, and low-affinity regions expose gaps in the recorded coverage.
    This supplies quantitative evidence toward completeness and representativeness requirements, but it does not establish every required aspect of data quality.
    \\
    \midrule

    DA-07 (SURK-DA-07)
                       &
    \enquote{The applicant should validate each of the requirements captured under Objectives DA-02, DA-03, DA-04, DA-05 and the architecture captured under Objective DA-06.}
                       &
    The similarity notion of \cref{def:ODDSimilarity} and the held-out evaluation in \cref{sec:Validation} quantify discrepancies between an independently specified ODD and the data-driven representation.
    This evidence can focus subject-matter-expert review, but it does not replace that review.
    \\
    \midrule

    DM-07 (SU-DM-06)
                       &
    \enquote{The applicant should ensure verification of the data, as appropriate, throughout the data management process so that the data management requirements (including the DQRs) are addressed.}
                       &
    Under the canonical ordering and tie rules, the construction is permutation-stable in exact arithmetic, making the ODD-derivation transformation reproducible.
    Unambiguous traceability to source and intermediate data remains the responsibility of the surrounding data-management process.
    \\
    \midrule

    DM-08 (SURK-DA-12, SU-DA-13)
                       &
    \enquote{The applicant should perform a data verification step to confirm the appropriateness of the defined ODD and of the data sets used for the training, validation and verification of the ML model.}
                       &
    \textbf{Optional, conditional support.}
    When independently specified ODD membership information is available, precision and recall quantify agreement with that reference, and held-out training, validation, or verification samples can be evaluated against the constructed support.
    Without independent reference membership, autoSAFE does not by itself establish the appropriateness of the final ODD or every aspect of dataset appropriateness.
    \\
    \midrule

    LM-07-SL (S-LM-04; SU-LM-05)
                       &
    \enquote{The applicant should account for the bias-variance trade-off in the model family selection and should provide evidence of the reproducibility of the model training process.}
                       &
    The RBF kernel family is fixed and justified in \cref{sec:Kernel}, and calibration reduces its remaining constants to \(\gamma{}\) and \(\lambda_{\mathrm{rel}}\) (cf.\ \cref{subsec:KernelParameters}).
    The ODD derivation has no random initialization or stochastic optimization and is reproducible for identical documented inputs; this supports the construction itself rather than the separate training of an AI/ML model.
    \\
    \midrule

    LM-11 (SU-LM-05)
                       &
    \enquote{The applicant should provide an analysis on the stability of the learning algorithms.}
                       &
    The derivation contains no stochastic model-training loop, so instability from random initialization or stochastic optimization does not apply.
    Permutation stability establishes repeatability for identical inputs, while the parameter and subset experiments characterize selected input changes; neither result proves stability under arbitrary perturbations, label errors, or distribution shift.
    \\
    \midrule

    LM-12 (SU-LM-12)
                       &
    \enquote{The applicant should perform and document the verification of the stability of the trained model, covering the whole AI/ML constituent ODD\@.}
                       &
    The explicit ODD representation and affinity can guide the selection of nominal, boundary, and low-coverage verification cases.
    autoSAFE does not evaluate the trained model's outputs under perturbations and therefore supplies only partial support for this objective.
    \\
    \midrule

    SA-02 (CRA-02)
                       &
    \enquote{The applicant should identify which data needs to be recorded for the purpose of supporting the continuous safety assessment.}
                       &
    Operational input vectors and their affinities are candidate records for later monitoring of ODD assumptions.
    The framework does not define data-integrity, retention, or retrieval mechanisms and does not establish distribution-shift detection.
    \\
    \midrule

    SA-03 (CRA-03)
                       &
    \enquote{In preparation of the continuous safety assessment, the applicant should define metrics, target values, thresholds and evaluation periods to guarantee that design assumptions hold.}
                       &
    The bounded affinity and membership threshold provide an ODD-membership metric and decision boundary that can contribute to monitoring.
    Safety target values, evaluation periods, and any relation between affinity and system-level safety-margin erosion require separate assurance decisions.
    \\
    \midrule

    EXP-19 (restructured in Issue 03 to: EXP-04, SA-05/RA-10)
                       &
    \enquote{Information concerning unsafe AI-based system operating conditions should be provided to the end user to enable them to take appropriate corrective action in a timely manner.}
                       &
    Affinity thresholds can supply an input to a separately designed warning or fallback function as operational support decreases.
    The affinity does not determine whether a condition is intrinsically unsafe, implement end-user communication, or guarantee timely corrective action.
    \\
    \bottomrule
\end{longtblr}

\end{document}